\pdfoutput=1

\documentclass[11pt]{article}

\usepackage[]{acl}

\usepackage{times}
\usepackage{latexsym}

\usepackage[T1]{fontenc}

\usepackage[utf8]{inputenc}
\usepackage{microtype}

\usepackage{amsmath}
\usepackage{graphicx}
\usepackage{booktabs}
\usepackage{multirow}
\usepackage[titletoc]{appendix}
\usepackage{booktabs}
\usepackage{subcaption}
\usepackage{xcolor}
\usepackage{tikz}
\usepackage{amsfonts}
\usepackage{tablefootnote}

\newcommand{\up}[1]{\textsuperscript{#1}}

\title{Discourse Structure Extraction from Pre-Trained and Fine-Tuned Language Models in Dialogues
} 

\author{Chuyuan Li\up{1}, Patrick Huber\up{2}, Wen Xiao\up{2}\\
    {\bf Maxime Amblard\up{1}, Chloé Braud\up{3}, Giuseppe Carenini\up{2}}\\
    \up{1} Université de Lorraine, CNRS, Inria, LORIA, F-54000 Nancy, France  \\
    \up{2} University of British Columbia, V6T 1Z4, Vancouver, BC, Canada \\
    \up{3} IRIT, Université de Toulouse, CNRS, ANITI, Toulouse, France\\
    \up{1}\texttt{\{firstname.name\}@loria.fr},     
    \up{3}\texttt{chloe.braud@irit.fr},\\
    \up{2}\texttt{\{huberpat, xiaowen3, carenini\}@cs.ubc.ca}
}

\begin{document}
\maketitle

\begin{abstract}
Discourse processing suffers from data sparsity, 
especially for dialogues. As a result, we explore approaches to 
build discourse structures for dialogues, based on attention matrices from Pre-trained Language Models (PLMs). We investigate multiple 
tasks for fine-tuning and show that the dialogue-tailored Sentence Ordering task performs best. 
To locate and exploit discourse information in PLMs, we propose an unsupervised and a semi-supervised method. Our proposals thereby achieve 
encouraging results on the 
STAC corpus, with F$_1$ scores of $57.2$ and $59.3$ for the unsupervised and semi-supervised methods, respectively. When restricted to projective trees, our scores improved to $63.3$ and $68.1$.


\end{abstract}

\section{Introduction}

In recent years,
the availability of accurate transcription methods and the increase in online communication have led to a vast rise 
in dialogue data, necessitating the development of automatic analysis systems.
For example, summarization of meetings or exchanges with customer service agents could be used to enhance collaborations or analyze customers issues 
\cite{li2019keep, feng2021survey};
machine reading comprehension in the form of question-answering could improve dialogue agents' performance and help knowledge graph construction \cite{he2021multi, li2021dadgraph}.
However, simple surface-level features are oftentimes not sufficient 
to extract valuable information from conversations
\cite{qin2017joint}. Rather, we need to understand the semantic and pragmatic relationships organizing the dialogue, for example through the use of discourse information.

\begin{figure}[]
    \centering
\vspace{0.5cm}
\includegraphics[width=.80\columnwidth]{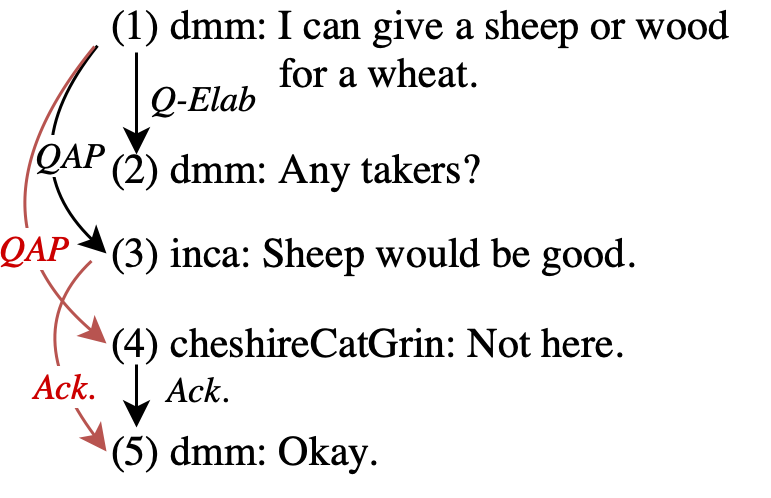}
    \caption{Excerpt of dependency structures in file \textit{s2-leagueM-game4}, STAC. 
    Red links are non-projective.
    }
    \label{fig:sdrt-example}
\vspace{-0.3cm}
\end{figure}

Along this line, several discourse frameworks have been proposed, 
underlying a variety of annotation projects. 
For dialogues, data has been primarily annotated within the Segmented Discourse Representation Theory (SDRT) \cite{asher2003logics}.
Discourse structures are thereby represented as dependency graphs with arcs linking spans of text and 
labeled with
semantico-pragmatic relations (e.g. \textit{Acknowledgment} (Ack) 
or \textit{Question-Answer Pair} (QAP)). Figure~\ref{fig:sdrt-example} shows an example  
from the Strategic Conversations corpus (STAC) \cite{asher2016discourse}.
Discourse processing refers to the 
retrieval of the inherit structure of coherent text,
and is often separated into three tasks: 
\texttt{EDU} segmentation, structure building (or attachment), and relation prediction.
In this work, we focus on the
automatic extraction of (naked) structures without discourse relations. 
This serves as a first critical step in creating a full discourse parser. It is important to note that 
naked 
structures have already been shown to be valuable features for specific tasks. 
\citet{louis2010discourse} mentioned that they are the most reliable indicator of importance in content selection. 
\citet{xu2020discourse,xiao2020we} on summarization, and \citet{jia2020multi}
on 
thread extraction, also demonstrated the advantages of 
naked structures.

Data sparsity has always been an issue for discourse parsing both in monologues and dialogues: 
the largest and most commonly used corpus annotated under the Rhetorical Structure Theory, the RST-DT \cite{carlson2003building} contains $21,789$ discourse units. 
In comparison, the largest dialogue discourse dataset (STAC) only contains 
$10,678$ units. 
Restricted to domain and size, the performance of supervised discourse parsers is still low, 
especially for dialogues, with at best $73.8\%$ F$_1$ for the naked structure on STAC \cite{wangstructure2021}.
As a result, several transfer learning approaches have 
been proposed, mainly focused on monologues. 
Previous work demonstrate 
that discourse information can be extracted from 
auxiliary tasks 
like sentiment analysis \cite{huber2020mega} and summarization \cite{xiao2021predicting}, 
or 
represented in 
language models \cite{koto2021discourse} and further enhanced by fine-tuning tasks \cite{huber2022plm4disc}.
Inspired by the latter approaches, 
we are pioneering in addressing this issue for dialogues and introducing effective semi-supervised and unsupervised strategies to uncover discourse information in large pre-trained language models (PLMs).
We find, however, that the monologue-inspired fine-tuning tasks 
are not performing well when applied to dialogues.
Dialogues are generally less structured, interspersed with more informal linguistic usage \cite{sacks1978simplest}, 
and have structural particularities
~\cite{asher2016discourse}.
Thus, we propose a new Sentence Ordering (SO) fine-tuning task 
tailored to dialogues. 
Building on the proposal in \citet{barzilay2008modeling}, we propose crucial, dialogue-specific extensions
with several novel shuffling strategies to enhance the pair-wise, inter-speech block, and inter-speaker discourse information in PLMs, and 
demonstrate its effectiveness 
over other fine-tuning tasks.

In addition, a key issue in using PLMs to extract document-level discourse information is how to choose the best attention head. 
We hypothesize that the location of discourse information in the network may vary, possibly influenced by the length and complexity of the dialogues. 
Therefore, we investigate methods that enables us to evaluate each attention head individually, 
in both unsupervised and semi-supervised settings. 
We introduce a new metric called ``Dependency Attention Support'' (DAS), which measures the level of support for the dependency trees generated by a specific self-attention head, allowing us to select the optimal head without any need for supervision.
We also propose a semi-supervised approach where a small validation set is used to choose the best head. 

Experimental results on the STAC dataset reveal that our unsupervised and semi-supervised methods outperform the strong LAST baseline (F$_1$ $56.8\%$, Sec.~\ref{sec:exp-setup}), delivering substantial gains 
on the complete STAC dataset (F$_1$ $59.3\%$, Sec.~\ref{subsec:semisup-result}) and show further improvements on the tree-structured subset (F$_1$ $68.1\%$, Sec.~\ref{subsec:proj-tree-analysis}).

To summarize, our contributions in this work are:
(1) Discourse information detection in pre-trained and sentence ordering fine-tuned LMs;
(2) Unsupervised and semi-supervised methods for discourse 
structure extraction from the attention matrices in PLMs; 
(3) Detailed quantitative and qualitative analysis of the extracted discourse structures.

\section{Related Work}

Discourse structures for complete documents have been mainly annotated within the Segmented Discourse Representation Theory (SDRT) \cite{asher2003logics} or the Rhetorical Structure Theory (RST) \cite{mann1988rhetorical}, with the latter leading to the largest corpora and many discourse parsers for monologues, while SDRT is the main theory for dialogue corpora, i.e., STAC \cite{asher2016discourse} and Molweni \cite{li2020molweni}. 
In SDRT, discourse structures are dependency graphs with possibly non-projective links (see Figure \ref{fig:sdrt-example}) 
compared to constituent trees structures in RST.
Early approaches to discourse parsing on STAC used varied decoding strategies, such as Maximum Spanning Tree algorithm 
\cite{muller2012constrained, li2014text, afantenos2012modelling} or Integer Linear Programming \cite{perret2016integer}. 
\citet{shi2019deep} first proposed a neural architecture based on hierarchical Gated Recurrent Unit (GRU) 
and reported $73.2$\% F$_1$ on STAC for naked structures.
Recently, \citet{wangstructure2021} adopted Graph Neural Networks (GNNs) and reported 
marginal improvements on the same test set
($73.8\%$ F$_1$).

Data sparsity being the issue, a new trend towards semi-supervised and unsupervised discourse parsing has emerged, almost exclusively for monologues. 
\citet{huber2019predicting, huber2020mega} leveraged sentiment information and showed promising results in cross-domain setting with the annotation of a silver-standard labeled corpus.
\citet{xiao2021predicting} extracted discourse trees from neural summarizers 
and confirmed the existence of discourse information in self-attention matrices.
Another line of work proposed to enlarge training data with a combination of several parsing models, as done in
\citet{jiang2016training, kobayashi2021improving, nishida2022out}.
In a fully unsupervised setting, \citet{kobayashi2019split} used similarity and dissimilarity scores for discourse tree creation, a method that can not be directly used for discourse graphs though.
As for dialogues, transfer learning approaches are rare.
\citet{badene2019data, badene2019weak} investigated a weak supervision paradigm where expert-composed heuristics,
combined to a generative model, 
are applied to unseen data.
Their method, however, 
requires domain-dependent annotation and a relatively large validation set for rule 
verification.
Another study by \citet{liu2021improving} focused on cross-domain transfer using STAC (conversation during online game) and Molweni (Ubuntu forum chat logs). 
They applied simple adaptation strategies (mainly lexical information) on a SOTA discourse parser and showed 
improvement 
compared to bare transfer: trained on Molweni and tested on STAC F$_1$ increased from $42.5\%$ to $50.5\%$.
Yet, their model failed to surpass simple baselines. 
Very recently, \citet{nishida2022out} investigated bootstrapping methods to adapt BERT-based parsers to out-of-domain
data with some success.
In comparison to all this previous work, to the best of our knowledge, we are the first to propose a fully unsupervised method and its extension to a 
semi-supervised setting.
 
As pre-trained language models 
such as BERT \cite{devlinbert}, BART \cite{lewis2020bart} or GPT-2 \cite{radford2019language} 
are becoming dominant in the field,
\textit{BERTology} research has gained much attention as an attempt 
to understand what kind of information these models capture. 
Probing tasks, for instance, can provide fine-grained analysis, 
but most of them only focus on sentence-level syntactic 
tasks \cite{jawahar2019does, hewitt2019structural, marevcek2019balustrades, kim2019pre, jiang2020can}.
As for discourse, 
\citet{zhu2020examining} and \citet{koto2021discourse} applied probing tasks and showed that BERT and BART encoders 
capture more discourse information than other models, like GPT-2.
Very recently, \citet{huber2022plm4disc} 
introduced a novel way to encode long documents and explored the effect of different fine-tuning tasks on PLMs, confirming that pre-trained and fine-tuned PLMs both can capture discourse information.
Inspired by  
these studies on monologues, we explore the use of PLMs to extract discourse structures in dialogues.

\section{Method: from Attention to Discourse}
\label{sec:method}

\begin{figure}
    \centering
    \includegraphics[width=0.98\columnwidth]{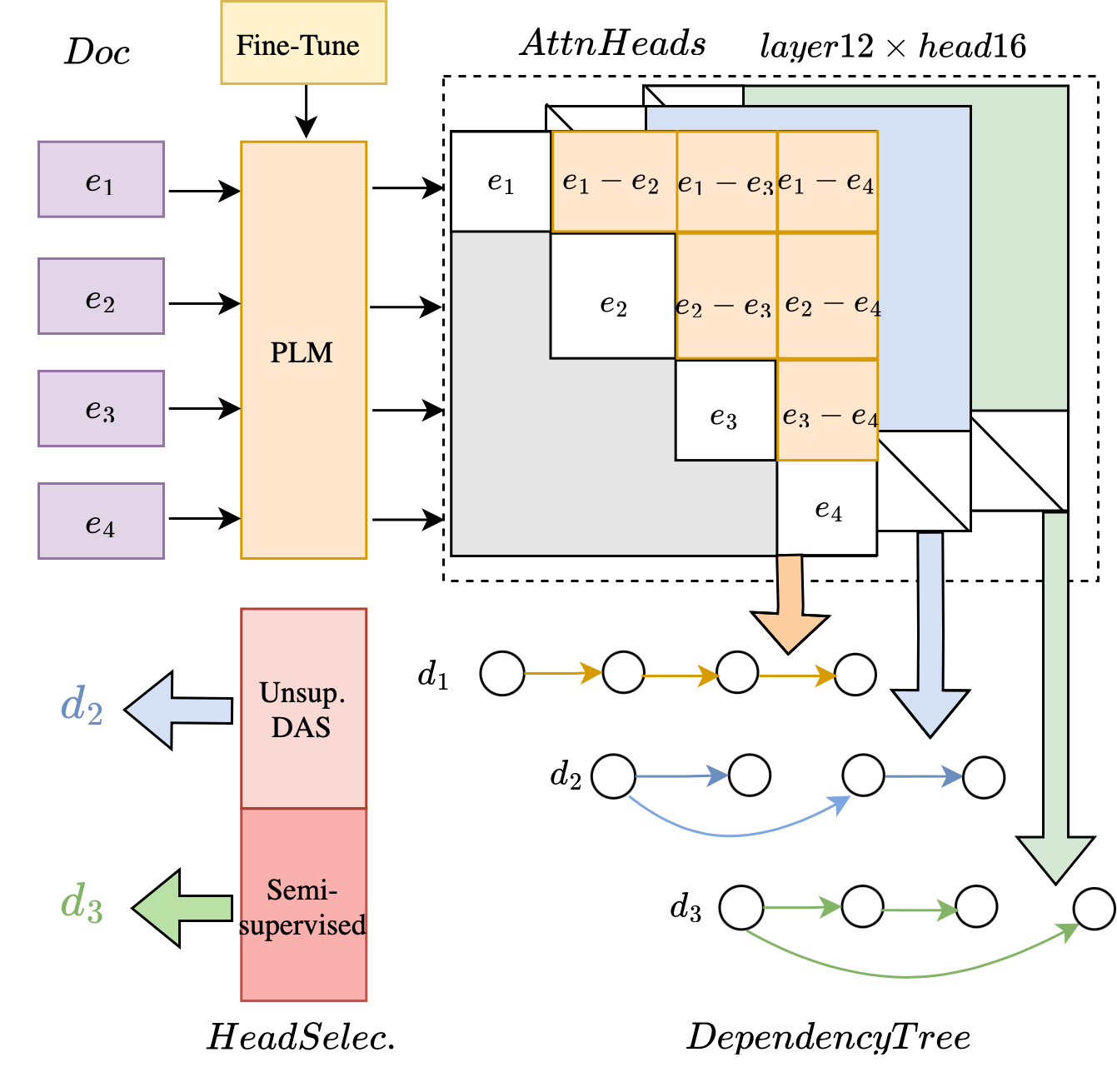}
    \caption{Pipeline for discourse structure extraction.}
    \label{fig:illustration-map}
\end{figure}

\subsection{Problem Formulation and Simplifications}
\label{subsec:method-problem-formulation}

Given a dialogue $D$ with $n$ \textit{Elementary Discourse Units} (\texttt{EDUs}) $\{e_1,e_2,e_3,...,e_n\}$, which are the minimal spans of text (mostly clauses, at most a sentence) to be linked by discourse relations,
the goal is to extract a Directed Acyclic Graph (DAG) 
connecting the $n$ \texttt{EDUs}
that best represents its SDRT discourse structure  
from attention matrices in PLMs\footnote{For more details on extracting discourse information from attention mechanisms see \citet{liu2018learning}.} (see Figure~\ref{fig:illustration-map} for an overview of the process). 
In our proposal, we make a few simplifications, partially adopted from previous work. We do not deal with SDRT \textit{Complex Discourse Units} (\texttt{CDUs}) 
following \citet{muller2012constrained} and \citet{afantenos2015discourse}, and do not tackle relation type assignment. Furthermore, similar to \citet{shi2019deep},
our solution can only generate discourse trees. 
Extending our algorithm to non-projective trees ($\approx6\%$ of edges are non-projectives in treelike examples) and graphs ($\approx 5\%$ of nodes with multiple incoming arcs) is left as future work.

\subsection{Which kinds of PLMs to use?}
\label{subsec:method-which-plm}

We explore both vanilla and fine-tuned PLMs, as they were both shown to contain discourse information for monologues \cite{huber2022plm4disc}.

\paragraph{Pre-Trained Models:}
We 
select BART \cite{lewis2020bart}, 
not only because its encoder has been shown to effectively capture discourse information,
but also because it dominated other alternatives in preliminary experiments, including DialoGPT \cite{zhang2020dialogpt} and DialogLM \cite{zhong2022dialoglm} - language models pre-trained with conversational data\footnote{See Appendix~\ref{append:other-plms} for additional results with other PLMs.}.

\paragraph{Fine-Tuning Tasks:}
We fine-tune BART on three discourse-related tasks: 

\textbf{(1) Summarization:} we use BART fine-tuned on the popular CNN-DailyMail (CNN-DM) news corpus~\cite{nallapati2016abstractive}, as well as on the SAMSum dialogue corpus~\cite{gliwa2019samsum}.

\textbf{(2) Question Answering:} we 
use BART fine-tuned on the latest version of 
the Stanford Question Answering Dataset (SQuAD 2.0) \cite{squad2}.

\begin{figure}
    \centering
    \includegraphics[width=\columnwidth]{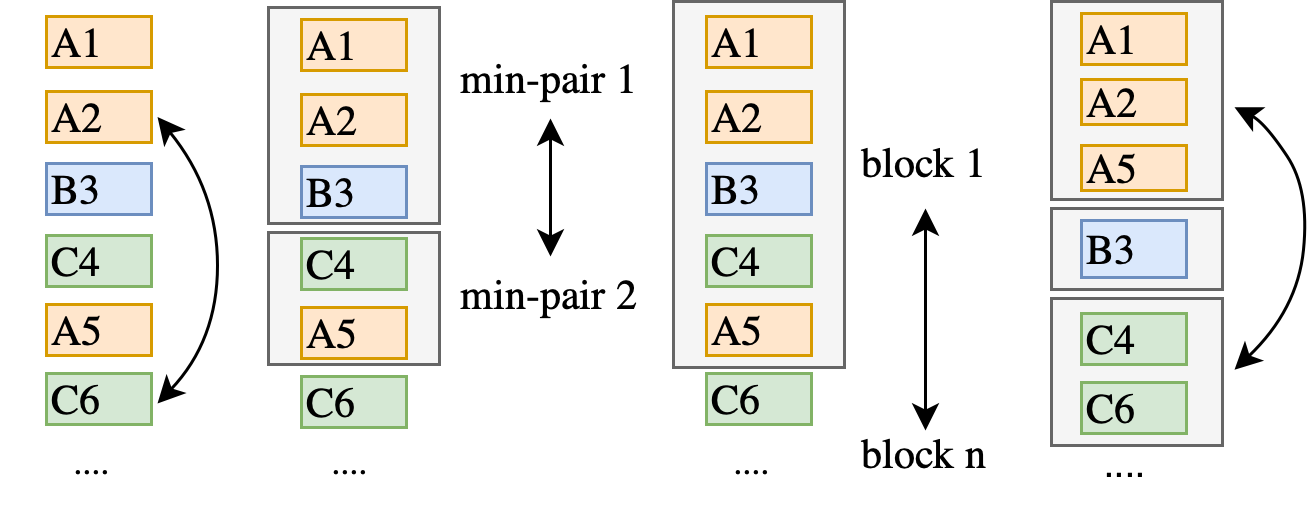}
    \caption{Shuffling strategies (left to right: partial, minimal-pair, block, speaker-turn) on a sequence of utterances 
    1 to 6, with A, B, C as the speakers.}
    \label{fig:shuffle}
\end{figure}

\textbf{(3) Sentence Ordering}: we fine-tune BART on the Sentence Ordering task -- reordering a set of shuffled sentences to their original order. 
We use an in-domain 
and an out-of-domain 
dialogue datasets (Sec.~\ref{sec:exp-setup}) for this task. 
Since fully random shuffling showed very limited improvements, we considered additional strategies to support a more gradual training tailored to dialogues.
Specifically, as shown in Figure~\ref{fig:shuffle}, we explore: (a) \textit{partial-shuf}: randomly picking $3$ utterances in a dialogue (or $2$ utterances if the dialogue is shorter than $4$) and shuffling them while maintaining the surrounding context.
(b) \textit{minimal-pair-shuf}: shuffling minimal pairs, comprising of a pair of speech turns from $2$ different speakers with at least $2$ utterances. A speech turn marks the start of a new speaker's turn in the dialogue.
(c) \textit{block-shuf}: shuffling a block containing multiple speech turns. We divide one dialogue into $[2,5]$ blocks based on the number of utterances\footnote{Block size is designed to be as twice or $3$ times bigger than ``min-pair'', we thus set criteria aiming to have $\approx 6$ \texttt{EDUs} per block: $|utt.| < 12: b=2$, $|utt.|\in[12, 22]: b=3$, $|utt.|\in[22, 33]: b=4$, $|utt.|\geq 33: n=5$.} and shuffle between blocks.
(d) \textit{speaker-turn-shuf}: grouping all speech productions of one speaker together. The sorting task consists of ordering speech turns from different speakers' production.
We evenly combine all permutations mentioned above to create our \textbf{mixed-shuf} data set and conduct the SO task. 

\paragraph{Choice of Attention Matrix:}
The BART model contains three kinds of attention matrices: encoder, decoder and cross attention. We use the encoder attention in this work, since it has been shown to capture most discourse information \cite{koto2021discourse} and outperformed the other alternatives in preliminary experiments on a validation set.

\subsection{How to derive trees from attention heads?}
\label{subsec:method-eisner}

Given an attention matrix $A^t\in \mathbb{R}^{k\times k}$ where $k$ is the number of tokens in the input dialogue, we derive the matrix $A^{edu}\in \mathbb{R}^{n\times n}$, 
with $n$ the number of \texttt{EDUs}, by computing $A^{edu}(i,j)$ as the average of the 
submatrix of $A^t$ corresponding to all the tokens of 
\texttt{EDUs} $e_i$ and $e_j$,
respectively. 
As a result, $A^{edu}$ captures how much \texttt{EDU} $e_i$ depends on \texttt{EDU} $e_j$ and can be used to generate a tree connecting 
all \texttt{EDUs} by maximizing their dependency strength.
Concretely, 
we find a Maximum Spanning Tree (MST) 
in the fully-connected dependency graph $A^{edu}$ using the Eisner algorithm \cite{eisner1996three}. 
Conveniently, since 
an utterance cannot be anaphorically and rhetorically
dependent on following utterances in a dialogue, 
as they are previously unknown \cite{afantenos2012modelling},
we can further simplify the inference by applying the following hard constraint to remove all backward links from the attention matrix $A^{edu}$:  
$a_{ij} = 0, \text{if } i>j$.

\subsection{How to find the best heads?}
\label{subsec:method-which-head}
\citet{xiao2021predicting} and \citet{huber2022plm4disc} showed that discourse information is not evenly distributed between heads and layers. However, they do not provide a strategy to select the head(s) containing most discourse information. Here, we propose two effective selection methods:  
fully unsupervised
or semi-supervised.

\subsubsection{Unsupervised Best Head(s) Selection}
\label{subsub:unsup}

\paragraph{Dependency Attention Support Measure (DAS):}
\label{para:method-as}
Loosely inspired by the confidence measure in \citet{nishida2022out}, 
where the authors define the confidence of a teacher model based on predictive probabilities of the decisions made,
we propose a DAS metric measuring the degree of support for the maximum spanning (dependency) tree (MST) from the attention matrix. 
Formally, given a dialogue $g$ with $n$ \texttt{EDUs}, we first derive the \texttt{EDU} matrix $A^{edu}$ from its attention matrix $A^g$ (see Sec.~\ref{subsec:method-eisner}). We then build the MST $T^g$ by selecting $n-1$ attention links $l_{ij}$ from $A^{edu}$ based on the tree generation algorithm.
DAS measures the strength of all those connections by computing the average score of all the selected links:
\begin{equation}
\label{as-measure}
    DAS(T^g)=\frac{1}{n-1} \sum_{i=1}^{n}\sum_{j=1}^{n} Sel(A^g,i,j)
\end{equation}
with $ Sel(A^g,i,j)=A^g_{ij}$, if $l_{ij}\in T^g$, $0$ otherwise.
Note that DAS can be easily adapted for a general graph by removing the restriction to $n-1$ arcs. 

\paragraph{Selection Strategy:} 
With DAS, we can now compute the degree of support from each attention head $h$ on each single example $g$ for the generated tree $DAS(T^g_h)$. We therefore propose two strategies to select attention heads based on the DAS measure, leveraging either global or local support.
The \textbf{global} support strategy selects the 
head with highest averaged DAS score over all the data examples:
\begin{equation}
\label{as-score-global}
H_{global} = \arg\max_{h} \sum_{g=1}^{M} DAS(T^g_h)
\end{equation}
where $M$ is the number of examples. 
In this way, we select the head that has a generally good performance on the target dataset. 

The second strategy is 
more adaptive to each document,
by only focusing on the \textbf{local} support. It does not select one specific head for the whole dataset, but instead selects the head/tree with the highest support for each single example $g$, i.e.,
\begin{equation}
\label{as-score-local}
    H_{local}^g = \arg\max_{h} DAS(T^g_h)
\end{equation}

\subsubsection{Semi-Supervised Best Head(s) Selection}
\label{subsub:semi-sup}
We also propose best heads selection using 
a few annotated examples. 
In conformity with real-world situations where labeled data is scarce, we sample three small subsets with $\{10, 30, 50\}$ data points (i.e., dialogues) from the validation set.
We examine every attention matrix individually, resulting in $12$ layers $\times$ $16$ heads candidate matrices
for each dialogue.
Then, the head with the highest micro-F$_1$ score on the 
validation set is selected to derive trees in the test set.
We also consider layer-wise aggregation, with details in Appendix~\ref{append:semisup-layer-wise-result}.

\section{Experimental Setup}
\label{sec:exp-setup}

\paragraph{Datasets: }
We evaluate our approach on predicting discourse dependency structures using the STAC corpus \cite{asher2016discourse}, a multi-party dialogue dataset annotated in the SDRT framework.
For the summarization and question-answering fine-tuning tasks, we use publicly available HuggingFace models \cite{wolf2020huggingface} (see Appendix~\ref{append:huggingface-models}). 
For the novel sentence ordering task, we train BART model on the STAC corpus and the DailyDialog corpus \cite{li2017dailydialog}. The key statistics for STAC and DailyDialog can be found in Table~\ref{tab:dataset}.
These datasets are split into train, validation, and test sets at $82\%$, $9\%$, $9\%$ and $85\%$, $8\%$, $8\%$ respectively. 
The Molweni corpus \cite{li2020molweni} is not included in our experiments due to quality issues, as detailed in Appendix~\ref{append:molweni}.

\begin{table}[]
\centering
\resizebox{\columnwidth}{!}{%
    \begin{tabular}{lrrrrr}
    \toprule
    Dataset & \#Doc & \#Utt/doc & \#Tok/doc & \#Spk/doc & Domain\\
    \midrule
    DailyDialog & $13,118$ & $13$ & $119$ & $2$ & Daily\\
     STAC & $1,161$ & $11$ & $50$ & $3$ & Game\\
    \bottomrule
    \end{tabular}}
    \caption{Key statistics of datasets. Utt = sentences in DD or \texttt{EDUs} in STAC; Tok = tokens; Spk = speakers.}
    \label{tab:dataset}
\end{table}

\paragraph{Baselines: }
We compare 
against the simple yet strong unsupervised 
LAST baseline \cite{schegloff2007sequence}, attaching every \texttt{EDU} to the previous one. 
Furthermore, to assess the gap between our approach and   
supervised dialogue discourse parsers, we 
compare with the Deep Sequential model by \citet{shi2019deep}
and the Structure Self-Aware (SSA) model by \citet{wangstructure2021}.


\paragraph{Metrics: }
We report the micro-F\textsubscript{1} and the Unlabeled Attachment Score (UAS) for the generated naked dependency structures.

\paragraph{Implementation Details: }
We base our work on the 
transformer implementations from the 
HuggingFace library \cite{wolf2020huggingface} 
and follow the \textit{text-to-marker} framework proposed in \citet{chowdhury2021everything} for the SO fine-tuning procedure. 
We use the original separation of train, validation, and test sets; 
set the learning rate to $5e-6$; 
use a batch size of $2$ for DailyDialog and $4$ for STAC, and train for $7$ epochs.
All other hyper-parameters are set following \citet{chowdhury2021everything}. We 
do not do any hyper-parameter tuning.
We omit $5$ documents in DailyDialog during training since the documents lengths exceed the token limit. We replace speaker names with markers (e.g. Sam $\rightarrow$ ``spk1''), following the preprocessing pipeline for dialogue utterances in PLMs.

\section{Results}
\label{sec:result}

\subsection{Results with Unsupervised Head Selection}
\label{subsec:unsup-result}


Results using our novel 
unsupervised DAS method on STAC are shown in  
Table~\ref{tab:plm-unsup-result} 
for both
the global (H\textsubscript{g}) and local (H\textsubscript{l}) head selection strategies.
These are compared to: (1) the unsupervised 
LAST baseline (at the top), which only predicts local attachments between adjacent \texttt{EDUs}. 
LAST is considered a strong baseline in discourse parsing \cite{muller2012constrained}, but has the obvious disadvantage of completely missing long-distance dependencies which may be critical in downstream tasks. 
(2) The supervised Deep Sequential parser by \citet{shi2019deep} and Structure Self-Aware model by \citet{wangstructure2021} (center of the table), both trained on STAC, reaching resp. $71.4\%$\footnote{We re-train the model, scores are slightly different due to different train-test splits, as in \citet{wangstructure2021}.} and $73.8\%$ in F$_1$.
 

In the last sub-table we show unsupervised scores from pre-trained and fine-tuned LMs on three auxiliary tasks: summarization, question-answering and sentence ordering (SO) with the mixed shuffling strategy. We present the global head (H\textsubscript{g}) and local heads (H\textsubscript{l}) performances selected by the DAS score (see section~\ref{subsub:unsup}). The best possible scores using an oracle head selector (H\textsubscript{ora}) are presented for reference.

Comparing the values in the bottom sub-table, we find that the pre-trained BART model 
under-performs LAST ($56.8$), with  global head and local heads achieving similar performance ($56.6$ and $56.4$ resp.).
Noticeably, models fine-tuned on the summarization task (``+CNN'', ``+SAMSum'') and question-answering (``+SQuAD2'') only add marginal improvements compared to BART. 
In the last two lines of the sub-table, we explore our novel sentence ordering fine-tuned BART models. We find that the BART+SO approach surpasses LAST when using local heads ($57.1$ and $57.2$ for DailyDialog and STAC resp.). 
As commonly the case, the intra-domain training performs best, which is further strengthened in this case due to the special vocabulary in STAC.
Importantly, 
our PLM-based unsupervised parser can capture some long-distance dependencies compared to LAST (Section~\ref{subsec:dep-tree-diversity}). 
Additional analysis regarding the chosen heads is in Section~\ref{subsec:as-measure-effect}.

\begin{table}[t!]
    \centering
    \resizebox{.9\columnwidth}{!}{%
    \begin{tabular}{lrrrr}
    \toprule
    Model\\
    \midrule
    \multicolumn{4}{l}{\textit{Unsupervised Baseline}} \\
    LAST &&& $56.8$\\
    \midrule
    \multicolumn{4}{l}{\textit{Supervised Models}} \\
    Deep-Sequential \shortcite{shi2019deep} &&& $71.4$ \\
    SSA-GNN \shortcite{wangstructure2021} &&& $73.8$ \\
    \midrule
    \textit{Unsupervised PLMs} & H$\textsubscript{g}$ & H$\textsubscript{l}$ & H$\textsubscript{ora}$\\
    BART & $56.6$ & $56.4$ & $57.6$\\
    \hspace{0.1 in}+ CNN & $56.8$ & $56.7$ & $57.1$\\
    \hspace{0.1 in}+ SAMSum & $56.7$ & $56.6$ & $57.6$\\
    \hspace{0.1 in}+ SQuAd2 & $55.9$ & $56.4$ & $57.7$\\
    \hspace{0.1 in}+ SO-DD & $56.8$ & $57.1$ & $58.2$\\
    \hspace{0.1 in}+ SO-STAC & $56.7$ & $\textbf{57.2}$ & $59.5$ \\
    \bottomrule
    \end{tabular}}
    \caption{Micro-F$_1$ on STAC for LAST, supervised SOTA models and unsupervised PLMs. 
    H$\textsubscript{g}$/H$\textsubscript{l}$/H$\textsubscript{ora}$: global/local/oracle heads. 
    Best (non-oracle) score in the $3^{rd}$ block is in bold. DD: DailyDialog.
    }
    \label{tab:plm-unsup-result}
\end{table}

\subsection{Results with Semi-Sup. Head Selection}
\label{subsec:semisup-result}
While the unsupervised strategy only delivered minimal improvements over the strong LAST baseline, 
Table~\ref{tab:semisup-result-stac} shows that if a few annotated examples are provided, it is possible to achieve substantial gains.
In particular, we report results on the vanilla BART model, as well as BART model fine-tuned on 
DailyDialog (``+SO-DD'') and STAC itself (``+SO-STAC'').
We execute $10$ runs for each semi-supervised setting ($[10, 30, 50]$) and report average scores and the standard deviation.



\begin{table}[]
    \centering
    \resizebox{\columnwidth}{!}{%
    \begin{tabular}{lrrr}
    \toprule
    Train on $\rightarrow$ & BART & + SO-DD & + SO-STAC\\
    Test with $\downarrow$ & F$_1$ & F$_1$ & F$_1$\\
    \midrule
    LAST BSL & $56.8$ & $56.8$ & $56.8$\\
    \midrule
    H$\textsubscript{ora}$ & $57.6$ & $58.2$ & $59.5$\\
    \midrule
    Unsup H$\textsubscript{g}$ & $\underline{56.6}$ & $56.8$ & $56.7$\\
    Unsup H$\textsubscript{l}$ & $56.4$ & $\underline{57.1}$ & $\underline{57.2}$\\
    \midrule
    Semi-sup $10$ & $57.0\textsubscript{0.012}$ & $57.2\textsubscript{0.012}$ & $57.1\textsubscript{0.026}$ \\
    Semi-sup $30$ & $57.3\textsubscript{0.005}$ & $57.3\textsubscript{0.013}$ & $59.2\textsubscript{0.009}$\\
    Semi-sup $50$ & $\textbf{57.4\textsubscript{0.004}}$ & $\textbf{57.7\textsubscript{0.005}}$ & $\textbf{59.3\textsubscript{0.007}}$\\
    \bottomrule
    \end{tabular}}
    \caption{Micro-F$_1$ on STAC from BART and SO fine-tuned BART with unsupervised and semi-supervised approaches. 
    Semi-supervised scores are averaged from $10$ random runs.
    Subscription is standard deviation.}
    \label{tab:semisup-result-stac}
\end{table}

The oracle heads (i.e., H$\textsubscript{ora}$) achieve superior performance compared to LAST. 
Furthermore, using a small scale validation set ($50$ examples) to select the best attention head 
remarkably improves the F$\textsubscript{1}$ score from $56.8\%$ (LAST) to $59.3\%$ (+SO-STAC).
F$\textsubscript{1}$ improvements across increasingly large validation-set sizes are consistent, accompanied by smaller standard deviations, as would be expected. 
The semi-supervised results are very encouraging: with $30$ annotated examples, we already reach a performance close to the oracle result, and with more examples we can further reduce the gap.

\section{Analysis}
\label{sec:analysis}
\subsection{Effectiveness of DAS}
\label{subsec:as-measure-effect}

\begin{figure}
    \centering
    \includegraphics[width=\columnwidth]{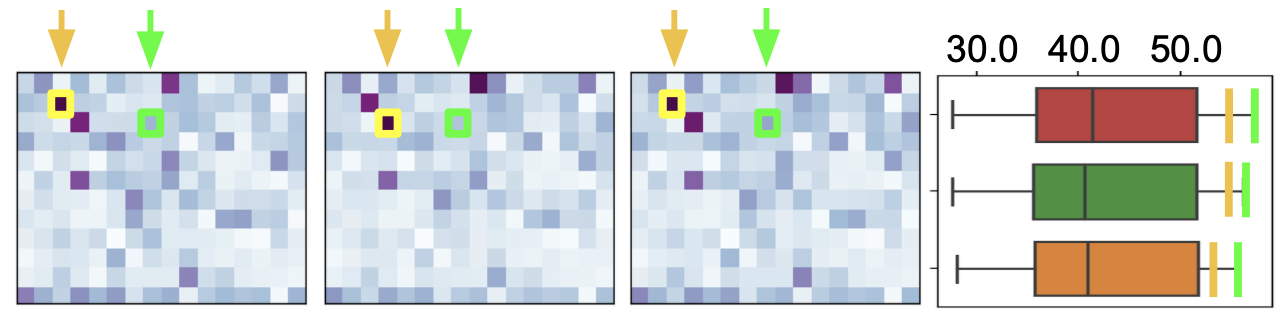}
    \caption{Heatmaps: DAS score matrices (layers: top to bottom=$12$ to $1$, heads: left to right=$1$ to $16$) for BART, BART+SO-DD, BART+SO-STAC. Darker purple=higher DAS score. \\
    Boxplot: Head-aggregated UAS scores for model BART (orange), BART+SO-DD (green) and BART+SO-STAC (red). Light green=head with highest UAS. Yellow=head with highest DAS score.}
    \label{fig:stac-box-heatmap}
\end{figure}

We now take a closer look at the performance degradation of our unsupervised approach based on DAS in comparison to the upper-bound defined by the performance of the oracle-picked head. 
To this end, Figure~\ref{fig:stac-box-heatmap} shows the DAS score matrices (left) for three models with the oracle heads and DAS selected heads highlighted in green and yellow, respectively.
These scores correspond to the global support strategy (i.e., H$\textsubscript{g}$).
It becomes clear that the oracle heads do not align with the DAS selected heads. Making a comparison between models, we find that discourse information is consistently located in deeper layers, with the oracle heads (light green) consistently situated in the same head for all three models. 
It is important to note that this information cannot be determined beforehand and can only be uncovered through a thorough examination of all attention heads.

While not aligning with the oracle, the top performing DAS heads (in yellow) are among the top $10\%$ best heads in all three models, as shown in the box-plot on the right. 
Hence, we confirm that the DAS method is a reasonable approximation to find discourse intense self-attention heads among the $12 \times 16$ attention matrices.

\subsection{Document and Arc Lengths}
\label{subsec:dep-tree-diversity}

The inherent drawback of the simple, yet effective LAST baseline is its inability to predict indirect arcs. To test if our approach can reasonably predict distant arcs of different length in the dependency trees, 
we analyze our results in regards to the arc lengths. Additionally, since longer documents tend to contain more distant arcs, 
we also examine the performance across different document lengths. 

\begin{figure}[b!]
    \centering
    \includegraphics[width=\columnwidth]{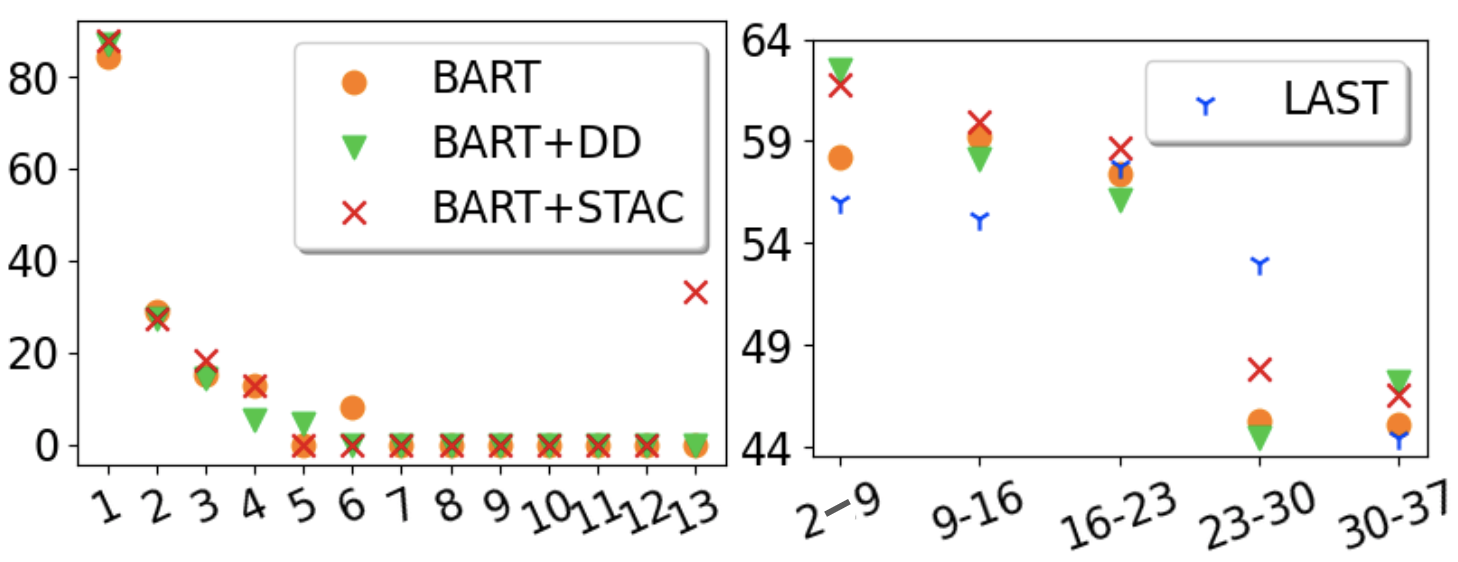}
    \caption{Left: UAS and arcs' distance. $x$ axis: arc distance. Right: averaged UAS for different length of document. $x$ axis: $\#$\texttt{EDUs} in a document. $y$ axis: UAS.}
    \label{fig:uas-doc-len-distant-arc}
\end{figure}

\paragraph{Arc Distance:}
To examine the extracted discourse structures for data sub-sets with specific arc lengths, we present the UAS score plotted against different arc lengths on the left side in Figure~\ref{fig:uas-doc-len-distant-arc}. 
Our analysis thereby shows that direct arcs achieve high UAS score ($>80\%$), independent of the model used. We further observe that the performance drops considerably for arcs of distance two and onwards, with almost all models failing to predict arcs longer than $6$.
BART+SO-STAC model correctly captures an arc of distance $13$.
Note that the presence for long-distance arcs ($\geq6$) is limited, accounting for less than $5\%$ of all arcs. 

\begin{figure}[]
    \centering
    \includegraphics[width=\columnwidth]
    {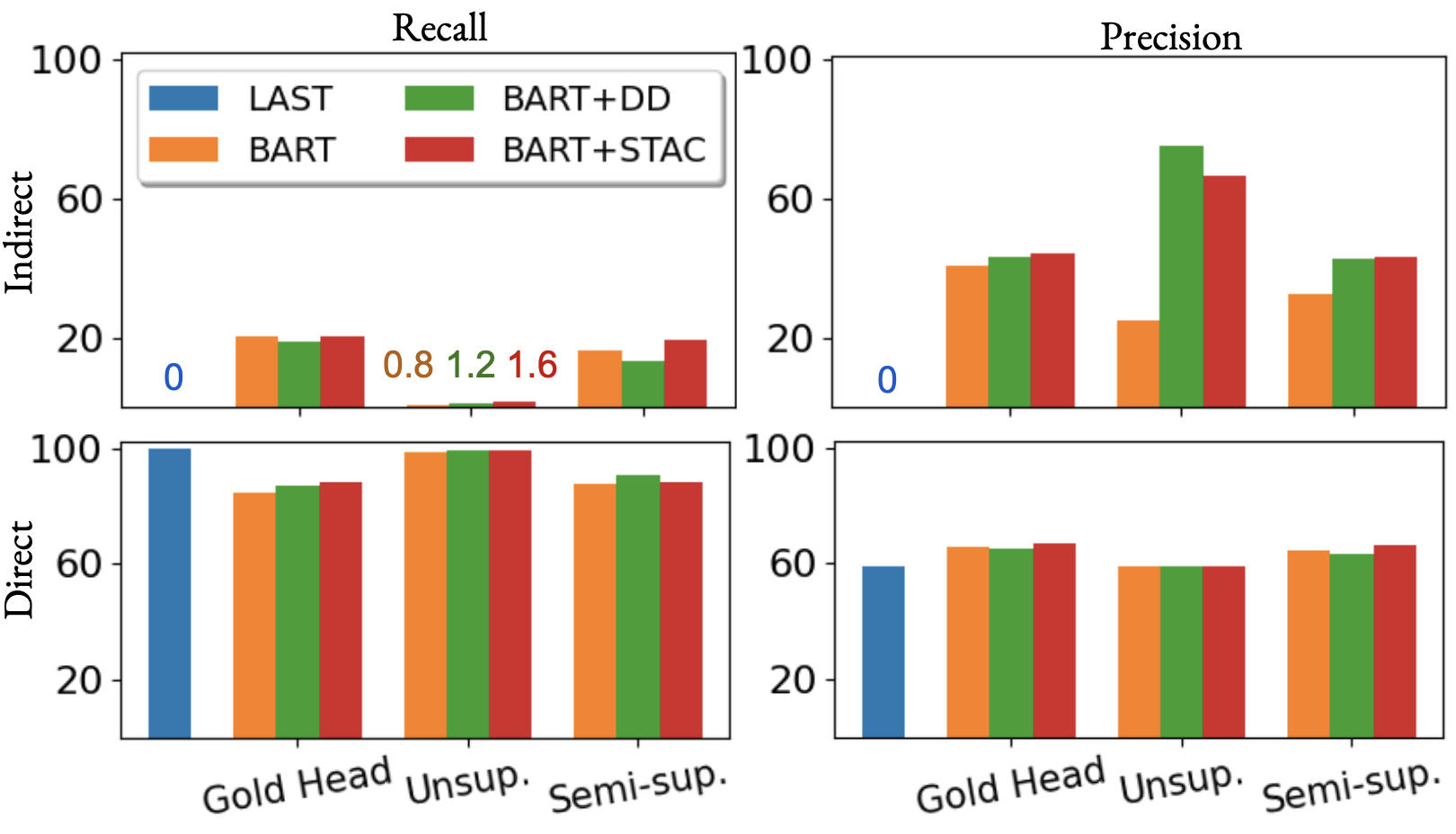}
    \caption{Comparison of recall (left) and precision (right) scores of indirect (top) and direct (bottom) links in LAST, BART, and SO fine-tuned BART models.}
    \label{fig:analyse-dist-dire-pre-rec}
\end{figure}

We further analyze the precision and recall scores when separating dependency links into \textit{direct} (adjacent forward arcs) and \textit{indirect} (all other non-adjacent arcs), following \citet{xiao2021predicting}.
For direct arcs, all models perform reasonably well (see Figure~\ref{fig:analyse-dist-dire-pre-rec} at the bottom). The precision is higher ($\approx$ $+6\%$ among all three BART models) and recall is lower than the baseline ($100\%$), indicating that our models predict less direct arcs but more precisely.
For indirect arcs (top in Figure~\ref{fig:analyse-dist-dire-pre-rec}), the best model is BART+SO-STAC ($20\%$ recall, $44\%$ precision), closely followed by original BART ($20\%$ recall, $41\%$ precision). In contrast, the LAST baseline model completely fails in this scenario ($0$ precision and recall).

\paragraph{Document Length:}
Longer documents tend to be more difficult to process because of the growing number of possible discourse parse trees. 
Hence, we analyze the UAS performance of documents in regards to their length, here defined as the number of \texttt{EDUs}. Results 
are presented on the right side in Figure~\ref{fig:uas-doc-len-distant-arc}, comparing the UAS scores for the three selected models and 
LAST for different document lengths. We split the document length range 
into $5$ even buckets between the shortest ($2$ \texttt{EDUs}) and longest ($37$ \texttt{EDUs}) document, resulting in 
$60$, $25$, $16$, $4$ and $4$ examples per bucket. 

For documents with less than $23$ \texttt{EDUs}, all fine-tuned models perform better than LAST, with BART fine-tuned on STAC reaching the best result. 
We note that PLMs exhibit an increased capability to predict distant arcs in longer documents. However, in the range of $[23, 30]$, the PLMs are inclined to predict a greater number of false positive distant arcs,
leading to under-performance compared to the LAST baseline.
As a result, we see that longer documents ($\geq23$) are indeed more difficult to predict, 
with even the performance of our best model (BART+STAC) strongly decreasing.


\subsection{Projective Trees Examination}
\label{subsec:proj-tree-analysis}
Given the fact that our method only extracts projective tree structures, we now conduct an additional analysis, exclusively examining the subset of STAC containing projective trees, 
on which our method could in theory achieve perfect accuracy.


\begin{table}[]
    \centering
    \resizebox{\columnwidth}{!}{
    \begin{tabular}{lrrrrr}
    \toprule
    & & \multicolumn{2}{c}{$\#$\texttt{EDUs}} & \multicolumn{2}{c}{$\#$Arcs}\\
    \cmidrule(lr){3-4} \cmidrule(lr){5-6}
    & $\#$Doc & Single-in & Multi-in & Proj. & N-proj.\\
    \midrule
    (1) Non-Tree & $48$ & $706$ & $79$ & $575$ & $170$\\
    (2) Tree & $61$ & $444$ & $0$ & $348$ & $35$ \\
    \hspace{0.2 in}- \textbf{Proj. tree} & $\textbf{48}$ & $314$ & $0$ & $266$ & $0$ \\
    \bottomrule
    \end{tabular}}
    \caption{STAC test set ground-truth tree and non-tree statistics. 
    ``Single-in'' and ``multi-in'' means \texttt{EDU} with single or multiple incoming arcs. 
    }
    \label{tab:analysis-trees-stats}
\end{table}

Table~\ref{tab:analysis-trees-stats} gives some statistics for this subset (``proj. tree'').
For the $48$ 
projective tree examples, the document length decreases from an average of $11$ to $7$ \texttt{EDUs}, however, still contains $\approx40\%$ indirect arcs, keeping the 
task difficulty comparable.
The scores for the extracted structures
are presented in Table~\ref{tab:analysis-subset-tree-result-stac48}. As shown, all three unsupervised models outperform LAST.
The best model is still BART fine-tuned on STAC, followed by the inter-domain fine-tuned +SO-DD and BART models. 
Using the 
semi-supervised approach, we see further improvement with the F$_1$ score reaching $68\%$ ($+6\%$ than LAST). Degradation for direct and indirect edges' precision and recall scores see Appendix~\ref{append:stac-treeset-precision-recall}.

\begin{table}[]
    \centering
    \resizebox{\columnwidth}{!}{%
    \begin{tabular}{lrrr}
    \toprule
    Train on $\rightarrow$ & BART & + SO-DD & + SO-STAC\\
    Test with $\downarrow$ & F$_1$ & F$_1$ & F$_1$\\
    \midrule
    LAST BSL & $62.0$ & $62.0$ & $62.0$\\
    \midrule
    H$\textsubscript{ora}$ & $64.8$ & $67.4$ & $68.6$\\
    \midrule
    Unsup H$_g$ & $\underline{62.5}$ & $62.5$ & $62.1$\\
    Unsup H$_l$ & $62.1$ & $\underline{62.9}$ & $\underline{63.3}$\\
    \midrule
    Semi-sup $10$ & $54.6\textsubscript{0.058}$ & $59.2\textsubscript{0.047}$ & $61.6\textsubscript{0.056}$\\
    Semi-sup $30$ & $60.3\textsubscript{0.047}$ & $60.3\textsubscript{0.044}$ & $65.6\textsubscript{0.043}$\\
    Semi-sup $50$ & $\textbf{64.8\textsubscript{0.000}}$ & $\textbf{66.3\textsubscript{0.023}}$ & $\textbf{68.1\textsubscript{0.014}}$\\
    \bottomrule
    \end{tabular}}
    \caption{Micro-F$_1$ scores on STAC projective tree subset with BART and SO fine-tuned BART models.
    }
    \label{tab:analysis-subset-tree-result-stac48}
\end{table}

Following \citet{ferracane2019evaluating}, we analyze key properties of the $48$ gold trees compared to our extracted structures using the 
semi-supervised method. 
To test the stability of the derived trees, we use three different seeds to generate the shuffled datasets to fine-tune BART. Table~\ref{tab:analysis-prop-trees-seeds} presents the averaged scores and the standard deviation of the trees.
In essence, while the extracted trees are generally ``thinner'' and ``taller'' than gold trees and contain slightly less branches, they are well aligned with gold discourse structures and do not contain ``vacuous'' trees, where all nodes are linked to one of the first two \texttt{EDUs}.
\begin{table}[]
    \centering
    \resizebox{\columnwidth}{!}{
    \begin{tabular}{lrrrr}
    \toprule
    & Avg.branch & Avg.height & $\%$leaf & Norm. arc \\
    \midrule
    GT & $1.67$ & $3.96$ & $0.46$ & $0.43$\\
    \midrule
    BART & $1.20$ & $5.31$ & $0.31$ & $0.34$\\
    \hspace{0.05 in}+SO-DD & $1.32\textsubscript{0.014}$ & $5.31\textsubscript{0.146}$ & $0.32\textsubscript{0.019}$ & $0.37\textsubscript{0.003}$\\
    \hspace{0.05 in}+SO-STAC & $1.27\textsubscript{0.076}$ & $5.28\textsubscript{0.052}$ & $0.32\textsubscript{0.011}$ & $0.35\textsubscript{0.015}$\\
    \bottomrule
    \end{tabular}}
    \caption{Statistics for gold and extracted projective trees 
    in BART and fine-tuned BART models.}
    \label{tab:analysis-prop-trees-seeds}
\end{table}
Further qualitative analysis of inferred structures is presented in Appendix~\ref{append:good-bad-examples}. 
Tellingly, on two STAC examples our model succeeds in predicting $>82\%$ of projective arcs, some of which span across $4$ \texttt{EDUs}.
This is encouraging, providing anecdotal evidence that our method is suitable to extract reasonable discourse structures. 

\subsection{Performance with Predicted \texttt{EDUs}}

Following previous work, all our experiments have started with gold-standard \texttt{EDU} annotations. However, this would not be possible in a deployed discourse parser for dialogues. To assess the performance of such system, we conduct additional experiments in which we first perform \texttt{EDU} segmentation and then feed the predicted \texttt{EDUs} to our methods.

To perform \texttt{EDU} segmentation, we employ the DisCoDisCo model \cite{gessler2021discodisco}, pre-trained on a random sample of 50 dialogues from the STAC validation set. We repeat this process three times to accommodate instability. Our results, as shown in Table~\ref{tab:edu-pred-stats}, align with those previously reported in \citet{gessler2021discodisco} ($94.9$), with an F-score of $94.8$. In the pre-training phase, we utilize all $12$ hand-crafted features\footnote{Such as POS tag, UD deprel, sentence length, etc..}, and opt for treebanked data for enhanced performance ($94.9$ compared to $91.9$ for plain text data). The treebanked data is obtained using the Stanza Toolkit \cite{qi2020stanza}.

\begin{table}[]
    \centering
    \resizebox{\columnwidth}{!}{%
    \begin{tabular}{rrrrr}
    \toprule
    Gold $\#$ & Predicted $\#$ & Precision $\%$ & Recall $\%$ & F$\textsubscript{1}$ $\%$ \\
    \midrule
    $1155$ & $1081$ & $96.0$ & $93.4$ & $94.8$ \\
    \bottomrule
    \end{tabular}}
    \caption{\texttt{EDU} segmentation results on STAC test set using DisCoDisCo model \cite{gessler2021discodisco}, which is re-trained on $50$ random dialogues from the validation set. Scores are averaged over three runs.
    }
    \label{tab:edu-pred-stats}
\end{table}

For evaluation, we adapt the discourse analysis pipeline proposed by \citet{joty2015codra}.
The results are shown in Table~\ref{tab:edu-result}, comparing the predicted and gold \texttt{EDUs}. The best head (i.e., H$\textsubscript{ora}$) performance decreases by $\approx 7$ points, from $59.5$ to $52.6$, as well as unsupervised and semi-supervised results. 
Despite the drop, our unsupervised and semi-supervised models still outperform the LAST baseline.
A similar loss of $\approx6$ points is also observed for RST-style parsing in monologues, as reported in \citet{nguyen2021rst}.

\begin{table}[]
    \centering
    \resizebox{\columnwidth}{!}{%
    \begin{tabular}{lrrrrrrr}
    \toprule
    & LAST & \multicolumn{3}{c}{Unsupervised} & \multicolumn{3}{c}{Semi-supervised} \\
    & & H$\textsubscript{g}$ & H$\textsubscript{l}$ & H$\textsubscript{ora}$ & semi-$10$ & semi-$30$ & semi-$50$\\
    \midrule
    Gold & $56.8$ & $56.7$ & $57.2$ & $59.5$ & $57.4\textsubscript{0.004}$ & $57.7\textsubscript{0.005}$ & $\textbf{59.3\textsubscript{0.007}}$\\
    Pred & $48.9$ & $50.8$ & $51.1$ & $52.6$ & $50.6\textsubscript{0.020}$ & $52.1\textsubscript{0.007}$ & $\textbf{52.2\textsubscript{0.004}}$\\
    \bottomrule
    \end{tabular}}
    \caption{Gold \texttt{EDUs} and predicted \texttt{EDUs} parsing results with BART+SO-STAC model. Scores for predicted \texttt{EDUs} are averaged over three runs.
    }
    \label{tab:edu-result}
\end{table}

\section{Conclusion}
\label{conclusion}
In this study, we explore approaches to build naked discourse structures from PLMs attention matrices to tackle the extreme data sparsity issue in dialogues. 
We show sentence ordering to be the best fine-tuning task and 
our unsupervised and semi-supervised methods for selecting the best attention head outperform a strong baseline, delivering substantial gains especially on tree structures.
Interestingly, discourse is consistently captured in deeper PLMs 
layers, and more accurately for shorter links. 

In the near future, we intend to explore graph-like structures from attention matrices, for instance, by extending treelike structures with additional arcs of high DAS score and applying linguistically motivated constraints, as in \citet{perret2016integer}. 
We would also like to expand shuffling strategies for sentence ordering and to explore other auxiliary tasks.
In the long term, our goal is to infer full discourse structures by incorporating the prediction of rhetorical relation types, all while remaining within unsupervised or semi-supervised settings.

\section*{Limitations}


Similarly to previous work, we have focused on generating only projective tree structures. This not only covers the large majority of the links ($\approx 94\%$), but it can also provide the backbone for accurately inferring the remaining non-projective links in future work.
We focus on the naked structure, as it is a significant first step and a requirement to further predict relations for discourse parsing.

We decided to run 
our experiments on the only existing high quality corpus, i.e., STAC. In essence, we traded-off generalizability for soundness of the results. A second corpus we considered, Molweni, had to be excluded due to serious 
quality issues.


Lastly, since we work with large language models and investigate every single attention head, 
computational efficiency is a concern.
We used a $4$-core GPU machine with the highest VRAM at 11MiB. The calculation for one discourse tree on one head was approximately $0.75$ seconds (in STAC the averaged dialogue length is $11$ \texttt{EDUs}), which quickly summed up to $4.5$ hours with only $100$ data points for $192$ candidate trees in one LM. 
When dealing with much longer documents, for example AMI and conversational section in GUM (in average $>200$ utterances/dialogue), our estimation shows that one dialogue takes up to $\approx 2$ minutes, which means $6.5$ hours for $192$ candidate trees.
Even though we use parallel computation, the exhaustive ``head'' computation results in a tremendous increase in time and running storage.
One possibility is to investigate only those ``discourse-rich'' heads, mainly in the deeper layers, for future work.

\section*{Ethical Considerations}


We carefully select the dialogue corpora used in this paper
to control for potential biases, hate-speech and inappropriate language by using human annotated corpora and professionally curated resources.
Further, we consider the privacy of dialogue partners in the selected datasets by replacing names with generic user tokens.

Since we are investigating the nature of the discourse structures captured in large PLMs, our work can be seen as making these models more transparent. This will hopefully contribute to avoid unintended negative effects, when the growing number of NLP applications relying on PLMs are deployed in practical settings.

In terms of environmental cost, the experiments described in the paper make use of Nvidia RTX 2080 Ti GPUs for tree extraction and Nvidia A100 GPUs for BART fine-tuning. We used up to $4$ GPUs for the parallel computation.
The experiments on corpus STAC took up to $1.2$ hours for one language model, and we tested a dozen models.
We note that while our work is based on exhaustive research on all the attention heads in PLMs to obtain valuable insights, future work will able to focus more on discourse-rich heads, which can help to avoid the quadratic growth of computation time for longer documents.

\section*{Acknowledgements}
The authors thank the anonymous reviewers for
their insightful comments and suggestions. This
work was supported by the PIA project “Lorraine
Université d’Excellence”, ANR-15-IDEX-04-LUE,
as well as the CPER LCHN (Contrat de Plan État-
Région - Langues, Connaissances et Humanités
Numériques). It was partially supported by the
ANR (ANR-19-PI3A-0004) through the AI Inter-
disciplinary Institute, ANITI, as a part of France’s
“Investing for the Future — PIA3” program, and
through the project AnDiAMO (ANR-21-CE23-
0020). It was also supported by the Language $\&$ Speech Innovation Lab of Cloud BU, Huawei Technologies Co., Ltd and the Natural Sciences and Engineering Research Council of Canada (NSERC). Nous remercions le Conseil de recherches en sciences naturelles et en génie du Canada (CRSNG) de son soutien.
Experiments presented were carried out
in clusters on the Grid’5000 testbed. We
would like to thank the Grid’5000 community
(https://www.grid5000.fr/).

\bibliography{biblio}
\bibliographystyle{acl_natbib}

\clearpage
\newpage
\appendix

\section{Semi-sup. Layer-Wise Results}
\label{append:semisup-layer-wise-result}

We consider both \textbf{layer-wise} attention matrices - averaging $16$ attention heads for every layer which gives $12$ candidate layers -, and \textbf{head-wise} attention matrices - taking each attention matrix individually which results in $192$ candidate matrices. 
Here we show results completed with layer-wise matrices for the whole test set and treelike examples in Table~\ref{tab:semisup-result-stac2} and Table~\ref{tab:analysis-subset-tree-result-stac48-2}.

\begin{table}[b!]
    \centering
    \resizebox{\columnwidth}{!}{%
    \begin{tabular}{lrrr}
    \toprule
    Train on $\rightarrow$ & BART & + SO-DD & + SO-STAC\\
    Test with $\downarrow$ & F$_1$ & F$_1$ & F$_1$\\
    \midrule
    Gold H & $57.6$ & $58.2$ & $59.5$\\
    \midrule
    Semi-sup-$10$ $1$L & $55.8\textsubscript{0.008}$ & $55.7\textsubscript{0.010}$ & $55.6\textsubscript{0.009}$\\
    Semi-sup-$30$ $1$L & $55.8\textsubscript{0.006}$ & $56.5\textsubscript{0.004}$ & $56.3\textsubscript{0.004}$\\
    Semi-sup-$50$ $1$L & $56.2\textsubscript{0.002}$ & $56.4\textsubscript{0.007}$ & $56.4\textsubscript{0.001}$\\
    Semi-sup-$10$ $1$H & $57.0\textsubscript{0.012}$ & $57.2\textsubscript{0.012}$ & $57.1\textsubscript{0.026}$ \\
    Semi-sup-$30$ $1$H & $57.3\textsubscript{0.005}$ & $57.3\textsubscript{0.013}$ & $59.2\textsubscript{0.009}$\\
    Semi-sup-$50$ $1$H & $\textbf{57.4\textsubscript{0.004}}$ & $\textbf{57.7\textsubscript{0.005}}$ & $\textbf{59.3\textsubscript{0.007}}$\\
    \bottomrule
    \end{tabular}}
    \caption{Micro-F$_1$ scores on STAC test set with BART and fine-tuned models. 
    H = ``head'', L = ``layer''. 
    Best semi-supervised score 
    is in bold. Subscription is std. deviation.}
    \label{tab:semisup-result-stac2}
\end{table}

\begin{table}[b!]
    \centering
    \resizebox{\columnwidth}{!}{%
    \begin{tabular}{lrrr}
    \toprule
    Train on $\rightarrow$ & BART & + SO-DD & + SO-STAC\\
    Test with $\downarrow$ & F$_1$ & F$_1$ & F$_1$\\
    \midrule
    Gold H & $64.8$ & $67.4$ & $68.6$\\
    \midrule
    Semi-sup-$10$ $1$L & $59.4\textsubscript{0.028}$ & $60.6\textsubscript{0.029}$ & $58.3\textsubscript{0.018}$\\
    Semi-sup-$30$ $1$L & $62.1\textsubscript{0.002}$ & $61.8\textsubscript{0.012}$ & $59.8\textsubscript{0.009}$\\
    Semi-sup-$50$ $1$L & $62.1\textsubscript{0.000}$ & $62.3\textsubscript{0.003}$ & $59.9\textsubscript{0.006}$\\
    Semi-sup-$10$ $1$H & $54.6\textsubscript{0.058}$ & $59.2\textsubscript{0.047}$ & $61.6\textsubscript{0.056}$\\
    Semi-sup-$30$ $1$H & $60.3\textsubscript{0.047}$ & $60.3\textsubscript{0.044}$ & $65.6\textsubscript{0.043}$\\
    Semi-sup-$50$ $1$H & $\textbf{64.8\textsubscript{0.000}}$ & $\textbf{66.3\textsubscript{0.023}}$ & $\textbf{68.1\textsubscript{0.014}}$\\
    \bottomrule
    \end{tabular}}
    \caption{Micro-F$_1$ scores on STAC projective tree subset with BART and SO fine-tuned BART models.
    }
    \label{tab:analysis-subset-tree-result-stac48-2}
\end{table}

\section{Molweni Corpus Quality Investigation}
\label{append:molweni}
\textbf{Molweni} \cite{li2020molweni} is a corpus derived from Ubuntu Chat Corpus \cite{lowe2015ubuntu}. It contains $10,000$ short dialogues between $8$ to $15$ utterances, annotated in SDRT framework. 

Considering the complexity of Ubuntu chat logs (multiple speakers, entangled discussion with various topics), we first conduct an examination of the corpus. 
Disappointingly, we found heavy repetition within sequential documents and inconsistency in discourse annotation among the same utterances. We thus decide not to include it in this work.

\begin{figure*}
    \centering
    \includegraphics[width=\columnwidth]{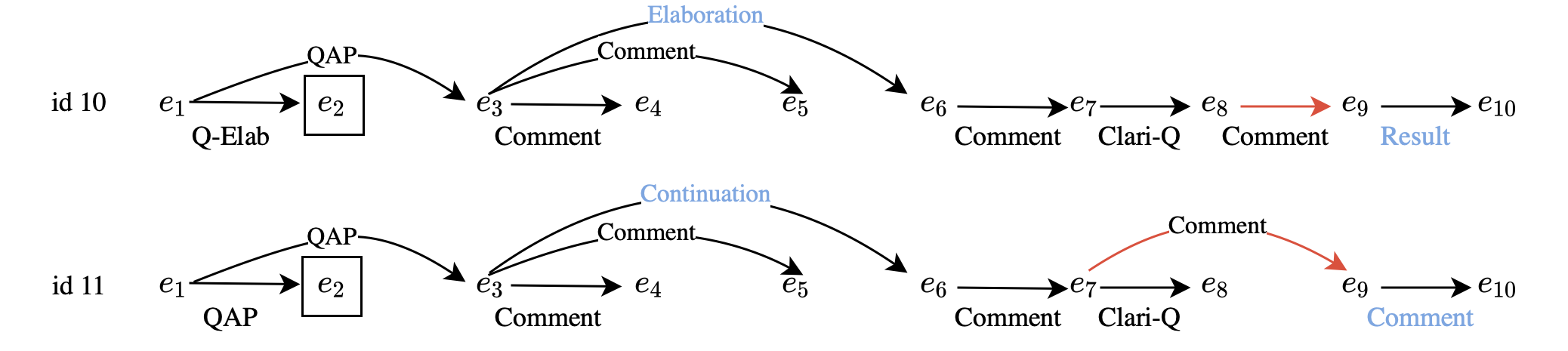}
    \caption{Similar documents in the same cluster. Circled \texttt{EDUs} are different. In red: inconsistent discourse arcs; in blue: inconsistent rhetorical relation.\\
    \textbf{test id 10}:\\
    $[e_1]$ matthew99857: so do i need additional hardware to fix it~?\\
    \textcolor{blue}{$[e_2]$ vocx: ca n't you disable the raid from the bios~? check your motherboard manual .}\\
    $[e_3]$ ikonia: just use the disk as an individual disk\\
    $[e_4]$ sugi: vocxi : oh i am sorry . i misunderstood you . thank i will try it now\\
    $[e_5]$ vocx: you need to word better your answers , seems like nobody in getting you today .\\
    $[e_6]$ sugi: vocx : iso 9660 cd-rom filesystem data udf filesystem data ( unknown version , id 'nsr01 ')\\
    $[e_7]$ ikonia: looks like that should work as a loop back file system\\
    $[e_8]$ sugi: -mount -o loop but instead of .iso .mdf ? or the .mds file ?\\
    $[e_9]$ ikonia: try it , linux see 's it as a `` image '' so it may work\\
    $[e_{10}]$ sugi: vocx : wow it worked , i feel retard for nto\\
    \textbf{test id: $11$}\\
    $[e_1]$ matthew99857: so do i need additional hardware to fix it~?\\
    \textcolor{blue}{$[e_2]$ ikonia: no you need to stop using raid}\\
    $[e_3]$ ikonia: just use the disk as an individual disk\\
    $[e_4]$ sugi: vocxi : oh i am sorry . i misunderstood you . thank i will try it now\\
    $[e_5]$ vocx: you need to word better your answers , seems like nobody in getting you today .\\
    $[e_6]$ sugi: vocx : iso 9660 cd-rom filesystem data udf filesystem data ( unknown version , id 'nsr01 ')\\
    $[e_7]$ ikonia: looks like that should work as a loop back file system\\
    $[e_8]$ sugi: -mount -o loop but instead of .iso .mdf ? or the .mds file ?\\
    $[e_9]$ ikonia: try it , linux see 's it as a `` image '' so it may work\\
    $[e_{10}]$ sugi: vocx : wow it worked , i feel retard for nto}
    \label{fig:invest-molweni}
\end{figure*}

\paragraph{Clusters:}
Among $500$ dialogues in discourse augmented test set, we found $105$ ``clusters''. 
One cluster groups all the documents with only one or two different utterances.
For instance, document id $10$ and $11$ are in the same cluster since only the second utterance is different (Figure~\ref{fig:invest-molweni}).
A similar situation is attested in the documents \{1, 2, 3\}, \{7, 8, 9\}, \{19, 20, 21\}, to name a few. 

\paragraph{Annotation Inconsistency:}
A closer examination of the annotation in similar examples reveals inconsistency for both discourse links and rhetorical relations. 
Precisely, we investigate every \textit{document pair} (two documents in the same cluster) in all $105$ clusters in the test set.
A visualization of inconsistency for documents $10$ and $11$ is shown in Figure~\ref{fig:invest-molweni}: apart from \texttt{EDU$_2$}, we expect the same links and relations among other \texttt{EDUs}. However, we observe one link inconsistency (in red) and two relation inconsistencies (in blue).
In total, we find $6\%$ of link errors ($\#$Err arc) within the same \texttt{EDUs} and $14\%$ of relation errors ($\#$Err rel) in the test set\footnote{For validation and train sets we find similar error rates.}. The scores are shown in Table~\ref{tab:mol-quanti}.

The Ubuntu Chat Corpus contains long dialogues with entangled discussion. A pre-processing had been made to generate shorter dialogues.
While these slightly different short dialogues could be interesting for other dialogue studies in the field. Our focus on the discourse structure request more various data points and most importantly, the coherent discourse annotation.

\begin{table}[]
    \centering
    \resizebox{\columnwidth}{!}{%
    \begin{tabular}{lrrrrr}
    \toprule
    Clus & Doc  & $\#$Theor & $\#$Err& $\#$Theor & $\#$Err\\
    ID & ID & $=$arc & arc & $=$rel & rel\\
    \midrule
    $1$ & \{1, 2, 3\}  & $18$ & $2$ & $16$ & $2$\\ 
    $2$ & \{7, 8, 9\}  & $18$ & $0$ & $18$ & $7$\\ 
    $3$ & \{10, 11, 12, 13, 14\} & $80$ & $4$ & $76$ & $25$\\ 
    ... & \\
    \midrule
    $105$ & $500$ & $4787$ & $284$ & $4503$ & $606$\\ 
    - & - & $100\%$ & $5.9\%$ & $100\%$ & $13.5\%$\\ 
    \bottomrule
    \end{tabular}}
    \caption{Quantitative resume of link and relation inconsistency in Molweni test set. ``Theor $=$arc'': number of arcs between the same utterances, \textit{a priori} should be linked in the same way; ``Theor $=$rel'': number of relations between the linked utterances.}
    \label{tab:mol-quanti}
\end{table}

\section{Precision and Recall Scores for Direct and Indirect Arcs in STAC Tree Set} 
\label{append:stac-treeset-precision-recall}

\begin{figure}
    \centering
    \includegraphics[width=\columnwidth]{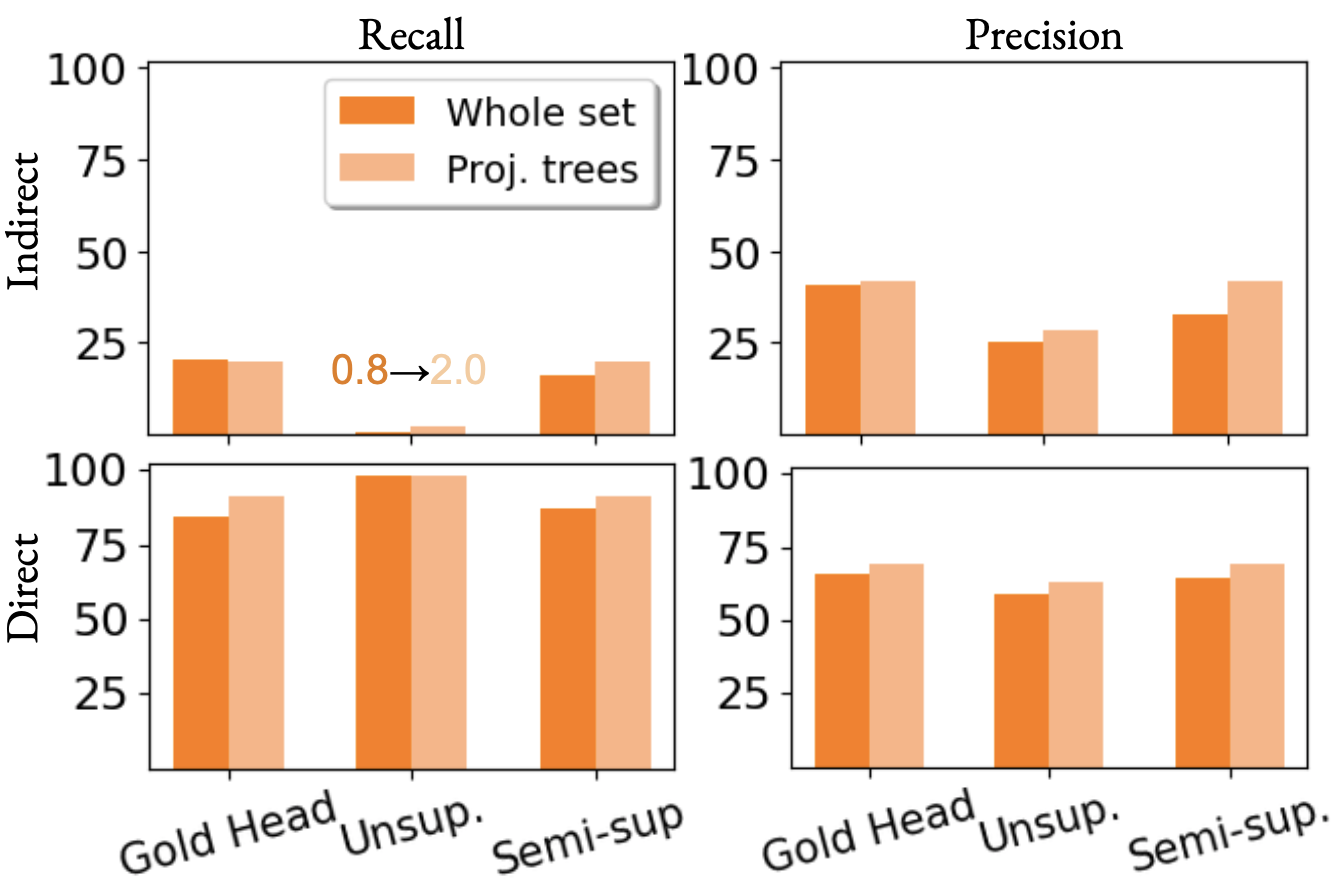}
    \caption{Recall and precision metrics in whole test set (darker color) \textit{vs.} projective tree subset (brighter color), with BART model.}
    \label{fig:stac-projtree-bart}
\end{figure}
\begin{figure}
    \centering
    \includegraphics[width=\columnwidth]{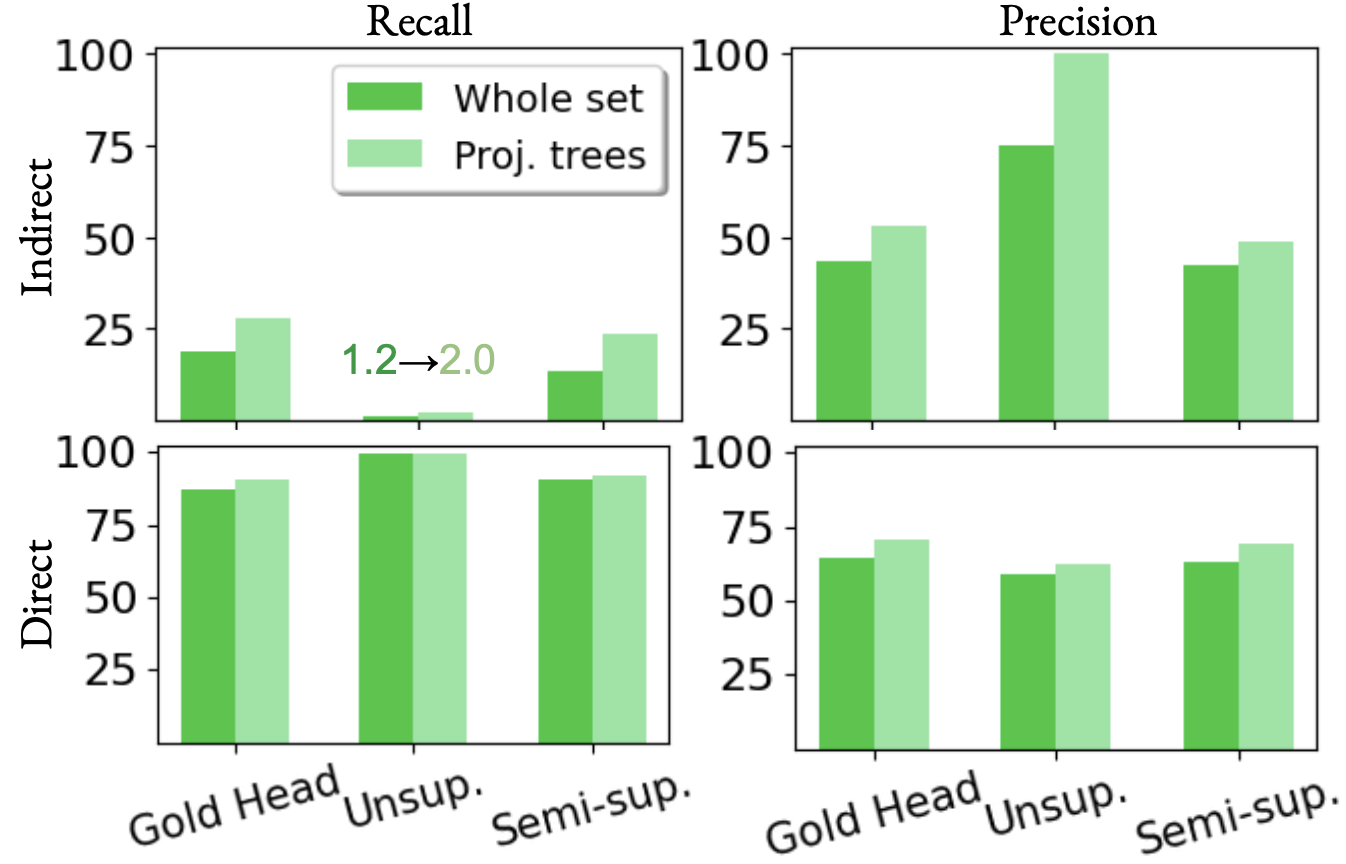}
    \caption{Recall and precision metrics in whole test set (darker color) \textit{vs.} projective tree subset (brighter color), with BART+SO-DD model.}
    \label{fig:stac-projtree-+dd}
\end{figure}
\begin{figure}
    \centering
    \includegraphics[width=\columnwidth]{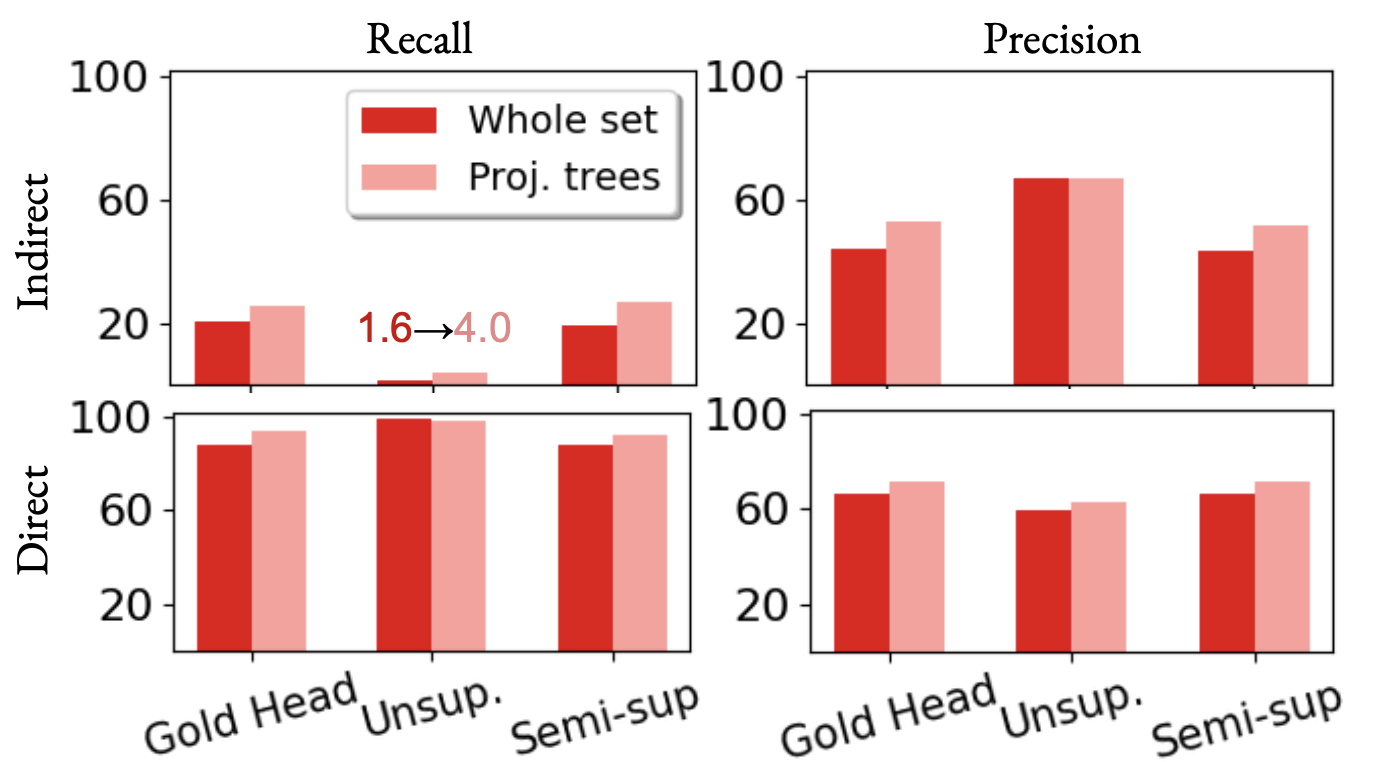} 
    \caption{Recall and precision metrics in whole test set (darker color) \textit{vs.} projective tree subset (brighter color), with model BART+SO-STAC.}
    \label{fig:stac-projtree-+stac}
\end{figure}

To compare the performance of the whole test set and tree-structured subset, we present the recall and precision scores of BART (Fig.~\ref{fig:stac-projtree-bart}), BART+SO-DD (Fig.~\ref{fig:stac-projtree-+dd}), and BART+SO-STAC (Fig.~\ref{fig:stac-projtree-+stac}) separately. 

\section{Qualitative Analysis in STAC}
\label{append:good-bad-examples}



\begin{figure*}
    \centering
    \includegraphics[width=\columnwidth]{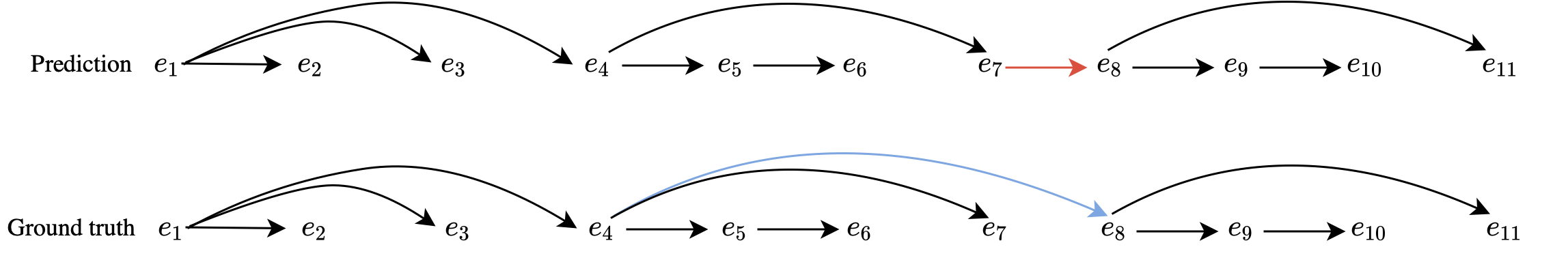}
    \caption{Well predicted example: \textit{pilot02-4}. $\#$\texttt{EDUs}: $11$. UAS: $90\%$. In red: FP arcs; in blue: FN arcs.\\
    $[e_1]$~Cat: anyone would give me clay? $[e_2]$~Thomas: none here $[e_3]$~william: no $[e_4]$~Cat: I have one wood to exchange $[e_5]$~Cat: any takers? $[e_6]$~william: no $[e_7]$~Cat: for sheep, wheat or clary $[e_8]$~Thomas: can I buy a sheep for two ore? $[e_9]$~william: have none $[e_{10}]$~Thomas: kk $[e_{11}]$~Cat: no sheep}
    \label{fig:quali-good1}
\end{figure*}

\begin{figure*}
    \centering
    \includegraphics[width=\columnwidth]{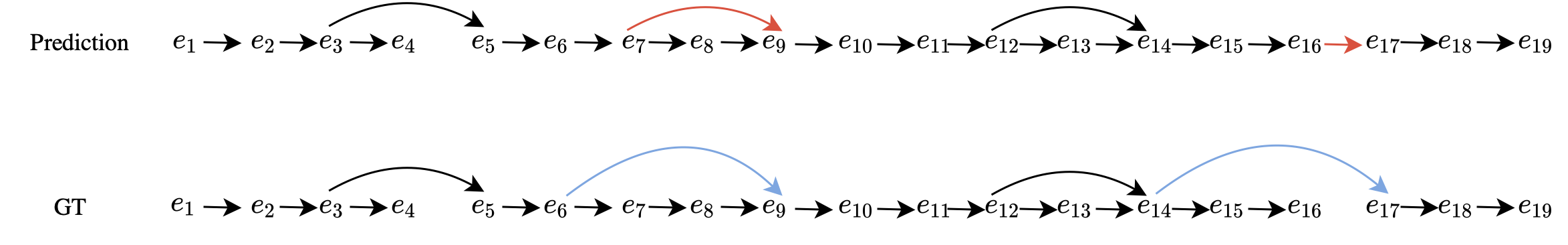}
    \caption{Well predicted example: \textit{pilot02-18}. $\#$\texttt{EDUs}: $19$. UAS: $88.9\%$. In red: FP arcs; in blue: FN arcs.\\
    $[e_1]$~william: hi markus. $[e_2]$~william: how many people are we waiting for? $[e_3]$~Thomas: think it's 1 more $[e_4]$~william: ok $[e_5]$~Markus: yes, one more $[e_6]$~Markus: seems there's a hickup logging into the game ... $[e_7]$~Thomas: that's ok, I not on a schedule $[e_8]$~Thomas: *I'm $[e_9]$~Markus: I guess you two had no problems joining the game? $[e_{10}]$~william: nope $[e_{11}]$~Markus: Ah great! $[e_{12}]$~Markus: So, one of you can now start the game. $[e_{13}]$~Markus: Have fun! $[e_{14}]$~william: the arrow is pointing at me $[e_{15}]$~william: but i cant press roll $[e_{16}]$~william: oh sorry $[e_{17}]$~Thomas: u can place a settlement $[e_{18}]$~Thomas: first $[e_{19}]$~Thomas: u roll later}
    \label{fig:quali-good2}
\end{figure*}

\begin{figure*}
    \centering
    \includegraphics{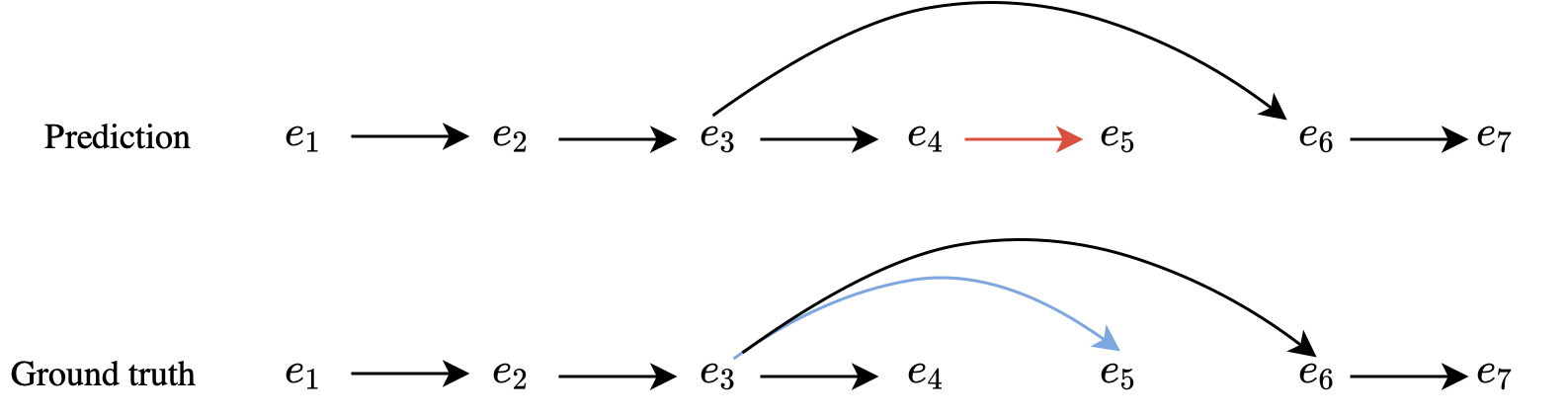}
    \caption{Well predicted example: \textit{s1-league3-game3}. $\#$\texttt{EDUs}: $7$. UAS: $83.3\%$. In red: FP arcs; in blue: FN arcs.\\
    $[e_1]$~Gaeilgeoir: ? $[e_2]$~yiin: build road $[e_3]$~inca: think we're meant to negotiate trades in the chat before offering $[e_4]$~yiin: oop $[e_5]$~yiin: ok then $[e_6]$~inca: part of the guys' experiment $[e_7]$~yiin: oh i see }
    \label{fig:quali-good3}
\end{figure*}

\begin{figure*}
    \centering
    \includegraphics{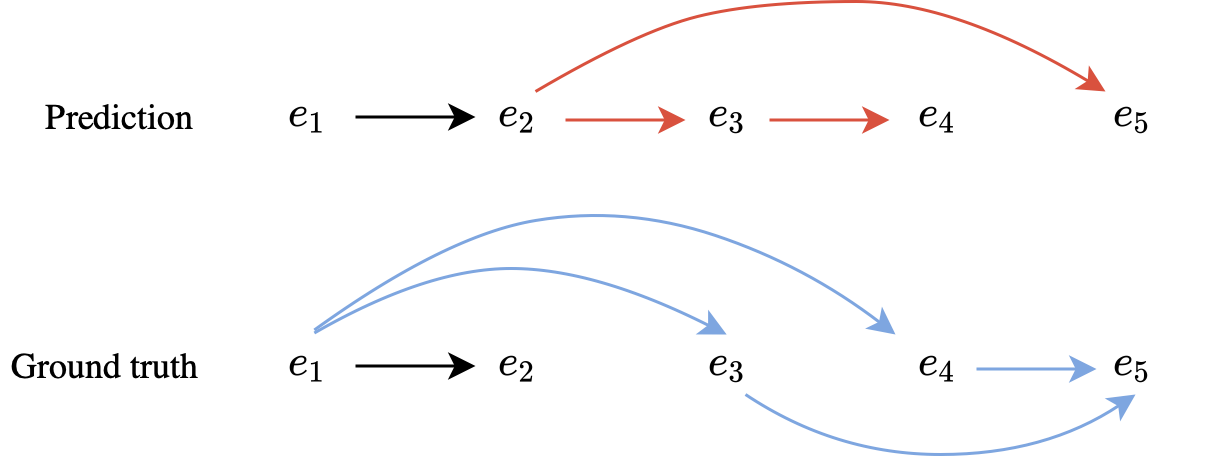}
    \caption{Badly predicted example: \textit{s2-leagueM-game4}. $\#$\texttt{EDUs}: $5$. UAS: $20\%$. In red: FP arcs; in blue: FN arcs.\\
    $[e_1]$~dmm: i can give a sheep or wood for a wheat. $[e_2]$~dmm: any takers? $[e_3]$~inca: sheep would be good. $[e_4]$~CheshireCatGrin: Not here. $[e_5]$~dmm: okay. 
    }
    \label{fig:quali-bad1}
\end{figure*}

\begin{figure*}
    \centering
    \includegraphics{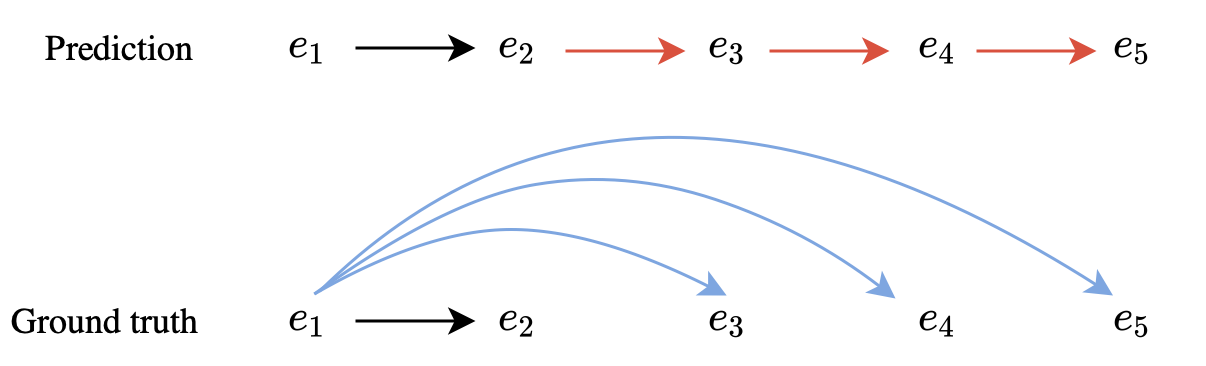}
    \caption{Badly predicted example: \textit{s1-league3-game3}. $\#$\texttt{EDUs}: $5$. UAS: $25\%$. In red: FP arcs; in blue: FN arcs.\\
    $[e_1]$~nareik15: anyone have ore. $[e_2]$~nareik15: I have some wood to trade. $[e_3]$~yiin: no sorry. $[e_4]$~inca: nope, sorry. $[e_5]$~Gaeilgeoir: no, sorry. 
    }
    \label{fig:quali-bad2}
\end{figure*}

\begin{figure*}
    \centering
    \includegraphics[width=\columnwidth]{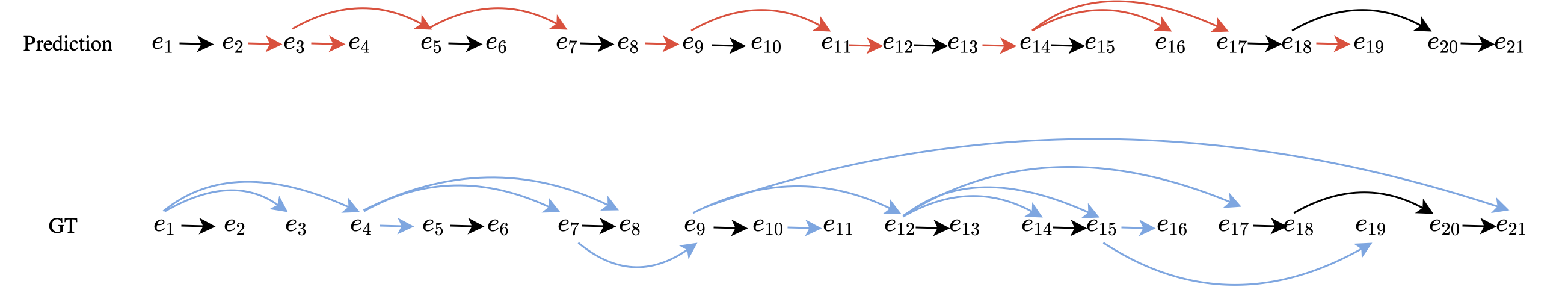}
    \caption{Badly predicted example: \textit{s1-league4-game2}. $\#$\texttt{EDUs}: $21$. UAS: $30\%$. In red: FP arcs; in blue: FN arcs.\\
    $[e_1]$~Shawnus: need wheat $[e_2]$~Shawnus: want..clay? $[e_3]$~ztime: you odo? $[e_4]$~ztime: yer.. $[e_5]$~ztime: I need clay.. $[e_6]$~ztime: can give wheat $[e_7]$~Shawnus: k $[e_8]$~Shawnus: this might be where i lose my road card a? $[e_9]$~ztime: er.. $[e_{10}]$~ztime: I think the trade is wrong? $[e_{11}]$~ztime: did you want wheat? $[e_{12}]$~Shawnus: yes $[e_{13}]$~Shawnus: for clay $[e_{14}]$~ztime: it said you wanted clay... $[e_{15}]$~somdechn: We all want wheat man $[e_{16}]$~somdechn: and clay... $[e_{17}]$~ztime: ok $[e_{18}]$~ztime: thanks.. $[e_{19}]$~Shawnus: haha $[e_{20}]$~Shawnus: thanks $[e_{21}]$~somdechn: That happens in the real game as well. 
    }
    \label{fig:quali-bad3}
\end{figure*}

\begin{figure*}
    \centering
    \includegraphics{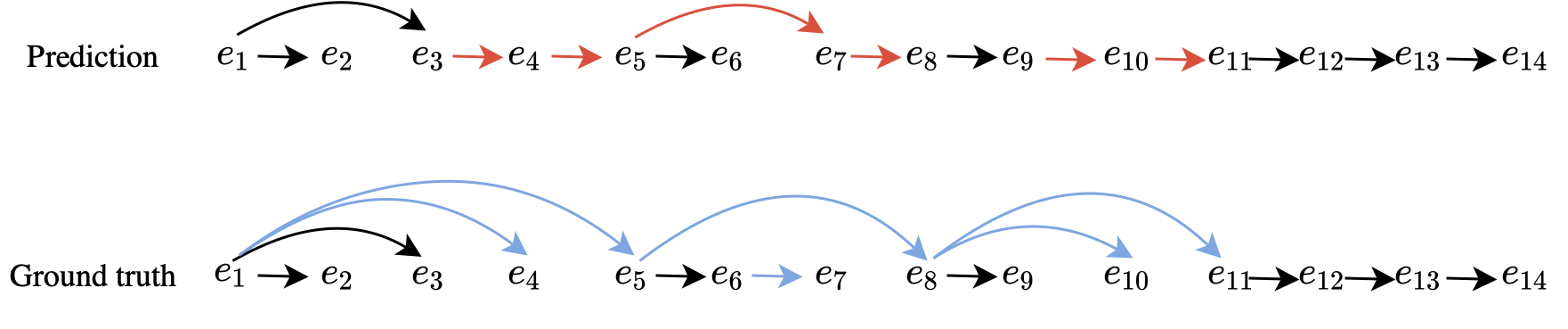}
    \caption{Random example: \textit{s2-league4-game2}. $\#$\texttt{EDUs}: $14$. UAS: $53.9\%$. In red: FP arcs; in blue: FN arcs.\\
    $[e_1]$~ztime: 7!!!! $[e_2]$~somdechn: Yeah right... $[e_3]$~ztime: what... is this a fix? $[e_4]$~Shawnus: hahaha $[e_5]$~ztime: ok anyone want wheat? $[e_6]$~Shawnus: nope $[e_7]$~Shawnus: just someone to roll 9's.. $[e_8]$~somdechn: Yes $[e_9]$~somdechn: I can give you wood. $[e_{10}]$~ztime: was that yes to a trade somdech? $[e_{11}]$~ztime: OK.. cool.. for 1 wheat? $[e_{12}]$~somdechn: and an ore.. :) $[e_{13}]$~ztime: err.. don't have ore.. $[e_{14}]$~ztime: thanks..}
    \label{fig:quali-random1}
\end{figure*}

\begin{figure*}
    \centering
    \includegraphics{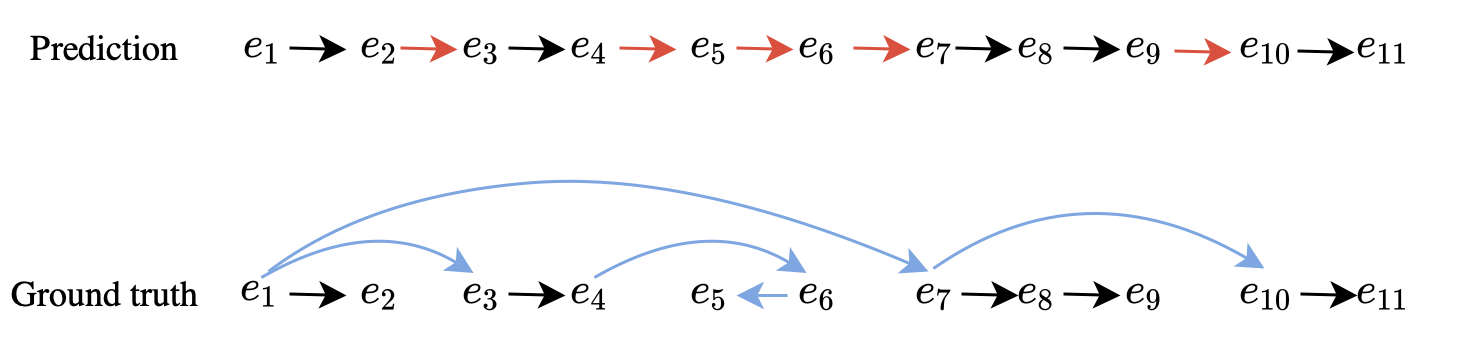}
    \caption{Random example: \textit{s1-league3-game3}. $\#$\texttt{EDUs}: $11$. UAS: $50\%$. In red: FP arcs; in blue: FN arcs.\\
    $[e_1]$~nareik15: anyone have wood to trade. I have sheep 
    $[e_2]$~yiin: no 
    $[e_3]$~Gaeilgeoir: Sorry, 
    $[e_4]$~Gaeilgeoir: I need wood too 
    $[e_5]$~Gaeilgeoir: I have wheat 
    $[e_6]$~Gaeilgeoir: if you want 
    $[e_7]$~inca: do you have wheat kieran? 
    $[e_8]$~inca: if so 
    $[e_9]$~inca: i can trade wood 
    $[e_{10}]$~nareik15: sorry, 
    $[e_{11}]$~nareik15: plenty of sheep though :) 
    }
    \label{fig:quali-random2}
\end{figure*}

We show a few concrete tree examples:
$3$ well predicted (Figure~\ref{fig:quali-good1}, \ref{fig:quali-good2}, \ref{fig:quali-good3}), $3$ badly predicted (Figure~\ref{fig:quali-bad1}, \ref{fig:quali-bad2}, \ref{fig:quali-bad3}), and $2$ random examples (Figure~\ref{fig:quali-random1}, \ref{fig:quali-random2}). 
Some patterns observed from badly predicted structures:
(1) chain-style prediction: as shown in Figure~\ref{fig:quali-bad2} and \ref{fig:quali-random2} where only adjacent \texttt{EDUs} are linked together; (2) inaccurate indirect arc prediction: especially for long documents such as the one in Figure~\ref{fig:quali-bad3}. 

\section{Results with other PLMs}
\label{append:other-plms}

We test with RoBERTa \cite{liu2019roberta}, DialoGPT \cite{zhang2020dialogpt}, and DialogLED (DialogLM with Longformer
) \cite{zhong2022dialoglm} to see how different language models encode discourse information. 
As shown in Table~\ref{tab:roberta-dialoglm-result}, the most discourse-rich head in RoBERTa slightly underperform BART ($-0.2\%$), so does the DialogLED ($-0.4\%$) and DialoGPT ($-1.4\%$).
Sentence ordering fine-tuned DialogLED model outperforms the original one, proving that our proposed SO task can help encoding the discourse information.

\begin{table}[t!]
    \centering
    \resizebox{\columnwidth}{!}{%
    \begin{tabular}{lr|rrrrr}
    \toprule
    Model &  & \multicolumn{2}{c}{Unsup} & \multicolumn{3}{c}{Semi-sup}\\
    & H$\textsubscript{ora}$ & H$\textsubscript{g}$ & H$\textsubscript{l}$  & Semi$10$ & Semi$30$ & Semi$50$\\
    \midrule
    BART & $57.6$ & $56.6$ & $56.4$ & $57.0\textsubscript{0.012}$ & $57.3\textsubscript{0.005}$ & $57.4\textsubscript{0.004}$\\
    \hspace{0.1 in}+ SO-DD & $58.2$ & $56.8$ & $57.1$ & $57.2\textsubscript{0.012}$ & $57.3\textsubscript{0.013}$ & $57.7\textsubscript{0.005}$\\
    \hspace{0.1 in}+ SO-STAC & $59.5$ & $56.7$ & $57.2$ & $57.1\textsubscript{0.026}$ & $59.2\textsubscript{0.009}$ & $\underline{\textbf{59.3}}\textsubscript{0.007}$\\
    \midrule
    RoBERTa & $57.4$ & $56.8$ & $56.8$ & $55.6\textsubscript{0.013}$ & $56.8\textsubscript{0.002}$ & $\underline{56.9}\textsubscript{0.003}$\\
    DialoGPT & $56.2$ & $42.7$ & $36.2$ & $52.9\textsubscript{0.043}$ & $55.1\textsubscript{0.017}$ & $\underline{56.2}\textsubscript{0.000}$\\
    DialogLED & $57.2$ & $56.8$ & $56.7$ & $54.6\textsubscript{0.026}$ & $54.7\textsubscript{0.061}$ & $\underline{56.6}\textsubscript{0.019}$\\
    \hspace{0.1 in}+ SO-DD & $57.7$ & $56.4$ & $56.6$ & $55.0\textsubscript{0.028}$ & $56.1\textsubscript{0.024}$ & $\underline{57.3}\textsubscript{0.009}$\\
    \hspace{0.1 in}+ SO-STAC & $58.4$ & $56.8$ & $57.1$ & $57.7\textsubscript{0.001}$ & $\underline{58.2}\textsubscript{0.005}$ & $57.7\textsubscript{0.001}$\\
    \bottomrule
    \end{tabular}}
    \caption{Micro-F$_1$ on STAC with other PLMs. Best score (except H$\textsubscript{ora}$) in each row is underlined.
    }
    \label{tab:roberta-dialoglm-result}
\end{table}

\section{Huggingface Models}
\label{append:huggingface-models}
Table~\ref{tab:huggingface-links} shows the models and the sources we obtained from Huggingface library \cite{wolf2020huggingface}.

\begin{table}[h]
    \centering
    \resizebox{\columnwidth}{!}{%
    \begin{tabular}{l}
    \toprule
    Model\\
    \midrule
    BART-large \\
    \url{https://huggingface.co/facebook/bart-large}\\
    BART-large-cnn\\
    \url{https://huggingface.co/facebook/bart-large-cnn}\\
    BART-large-samsum \\  \url{https://huggingface.co/linydub/bart-large-samsum}\\
    BART-large-finetuned-squad2\\  \url{https://huggingface.co/phiyodr/bart-large-finetuned-squad2}\\
    RoBERTa-large\\
    \url{https://huggingface.co/roberta-large}\\
    DialoGPT-small\\
    \url{https://huggingface.co/microsoft/DialoGPT-small}\\
    DialogLED-large-5120\\
    \url{https://huggingface.co/MingZhong/DialogLED-large-5120}\\
    \bottomrule
    \end{tabular}}
    \caption{Huggingface models and URLs.}
    \label{tab:huggingface-links}
\end{table}

\onecolumn
\clearpage
\newpage



\end{document}